\pdfoutput=1

\documentclass[11pt]{article}

\usepackage[final]{acl}

\usepackage{times}
\usepackage{latexsym}
\usepackage{hyperref}       
\usepackage{url}            
\usepackage{booktabs}       
\usepackage{amsfonts}       
\usepackage{nicefrac}       
\usepackage{microtype}      
\usepackage{xcolor}         
\usepackage{multirow}
\usepackage{graphicx}
\usepackage{bbding}
\usepackage{makecell}
\usepackage{amsthm,amsmath,amssymb}
\usepackage{adjustbox}
\usepackage{pifont}
\usepackage{subfigure}
\usepackage{tcolorbox}
\usepackage{parskip}
\usepackage{colortbl}

\definecolor{my_green}{RGB}{51,102,0}
\definecolor{my_red}{RGB}{204, 0, 0}
\renewcommand{\checkmark}{\textcolor{my_green}{\ding{51}}} 
\newcommand{\crossmark}{\textcolor{my_red}{\ding{55}}} 

\usepackage[T1]{fontenc}

\usepackage[utf8]{inputenc}

\usepackage{microtype}

\usepackage{inconsolata}

\title{TempCompass: Do Video LLMs Really Understand Videos?}

\newcommand*{\affaddr}[1]{#1}
\newcommand*{\affmark}[1][*]{\textsuperscript{#1}}
\author{
  Yuanxin Liu\affmark[\S]\thanks{Equal contribution.}, 
  Shicheng Li\affmark[\S]\footnotemark[1],
  Yi Liu\affmark[\S],
  Yuxiang Wang\affmark[\S],\\
  \textbf{Shuhuai Ren\affmark[\S]},
  \textbf{Lei Li\affmark[\dag]},
  \textbf{Sishuo Chen\affmark[\P],
  Xu Sun\affmark[\S],
  Lu Hou\affmark[\ddag]} \\
  \affaddr{\affmark[\S] National Key Laboratory for Multimedia Information Processing,\\
      School of Computer Science, Peking University}\\
  \affaddr{\affmark[\P] Center for Data Science, Peking University} \\
  \affaddr{\affmark[\dag] The University of Hong Kong} \quad \affaddr{\affmark[\ddag] Huawei Noah's Ark Lab} \\
  \texttt{\small\{liuyuanxin, yuxiangwang, shuhuai\_ren\}@stu.pku.edu.cn} \quad \texttt{\small{nlp.lilei@gmail.com}}\\
  \texttt{\small\{lisc99, imliuyi, chensishuo, xusun\}@pku.edu.cn} \quad \texttt{\small{houlu3@huawei.com}} \\
}

\begin{document}
\maketitle
\begin{abstract}
Recently, there is a surge in interest surrounding video large language models (Video LLMs). However, existing benchmarks fail to provide a comprehensive feedback on the temporal perception ability of Video LLMs. On the one hand, most of them are unable to distinguish between different temporal aspects (e.g., speed, direction) and thus cannot reflect the nuanced performance on these specific aspects. On the other hand, they are limited in the diversity of task formats (e.g., only multi-choice QA), which hinders the understanding of how temporal perception performance may vary across different types of tasks. Motivated by these two problems, we propose the \textbf{TempCompass} benchmark, which introduces a diversity of temporal aspects and task formats. To collect high-quality test data, we devise two novel strategies: (1) In video collection, we construct conflicting videos that share the same static content but differ in a specific temporal aspect, which prevents Video LLMs from leveraging single-frame bias or language priors. (2) To collect the task instructions, we propose a paradigm where humans first annotate meta-information for a video and then an LLM generates the instruction. We also design an LLM-based approach to automatically and accurately evaluate the responses from Video LLMs. Based on TempCompass, we comprehensively evaluate 8 state-of-the-art (SOTA) Video LLMs and 3 Image LLMs, and reveal the discerning fact that these models exhibit notably poor temporal perception ability. The data and evaluation code are available at \url{https://github.com/llyx97/TempCompass}.

\end{abstract}

\section{Introduction}
The development of video understanding systems has long been a popular topic in artificial intelligence research. Inspired by the unprecedented progress of large language models (LLMs), a line of initial efforts \citep{VideoChat,Video-LLaMA,Video-ChatGPT,Valley,Ren2023TimeChat} have been devoted to build LLMs with video understanding ability. These Video LLMs can serve as versatile multi-modal solvers for video and language tasks, demonstrating strong potential across various real-world applications.

With the rapid development of Video LLMs, a compelling question arises: ``Do Video LLMs really understand the temporal dynamics of videos'' Despite the importance of this question, current benchmarks fail to provide a satisfactory answer. \textbf{Firstly}, a majority of them neglect differentiating between various temporal aspects (e.g., type of action, speed and direction), thereby failing to offer a comprehensive view to diagnose the temporal perception ability. \textbf{Secondly}, while some Video LLM benchmarks \citep{AutoEval-Video,MVBench} have categorized various temporal aspects, they are restricted in task format variety (e.g., only multi-choice QA). Consequently, they are not optimally suited for assessing Video LLMs, which are expected to generalize across diverse tasks and instruction formats.

\begin{figure*}[t]
\centering
\includegraphics[width=1.\textwidth]{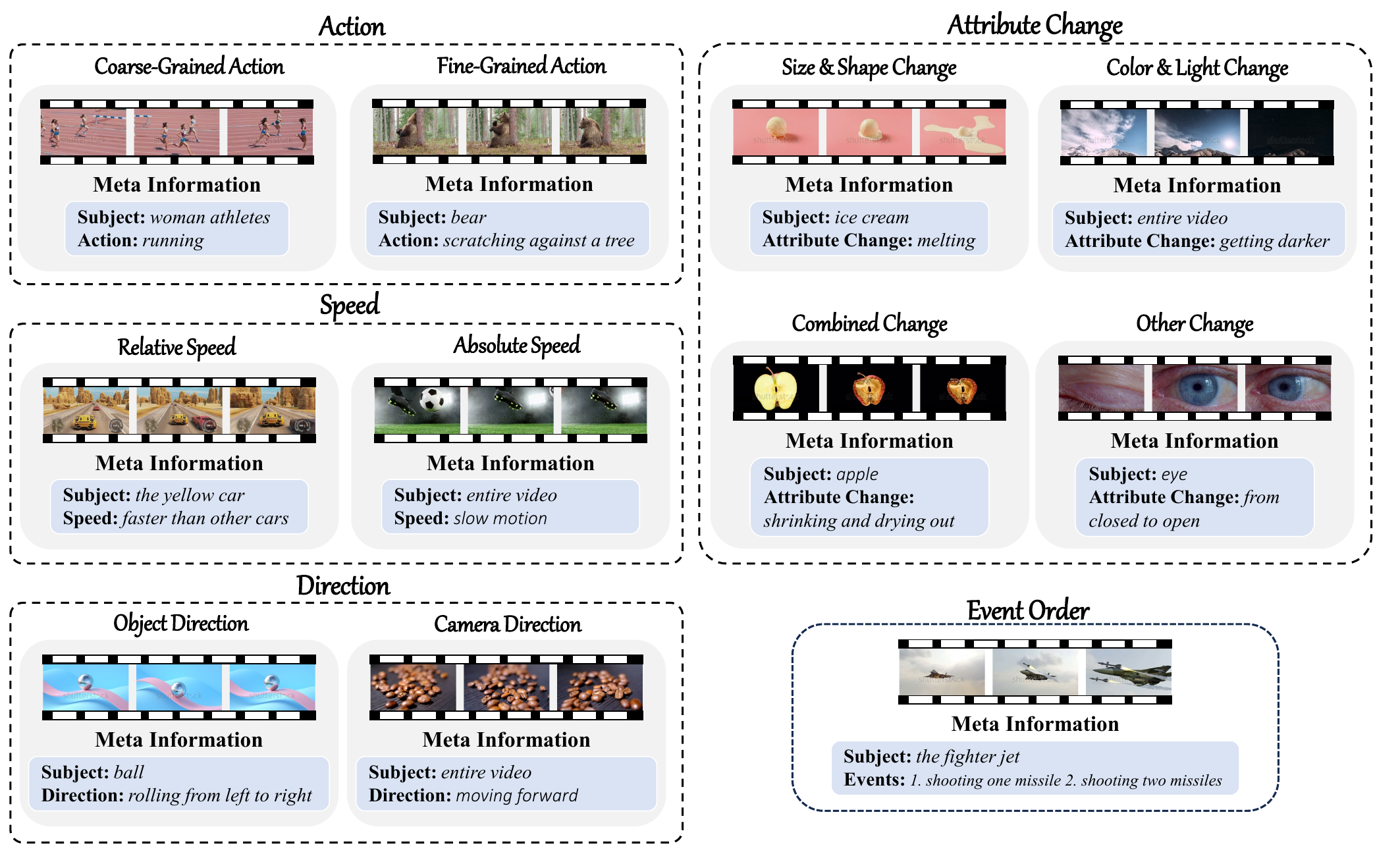}
\caption{Illustration of the temporal aspects (Section \ref{sec:temporal_aspects}) and meta-information (Section \ref{sec:meta_collect}).}
\label{fig:temporal_aspects}
\end{figure*}

In response to the above issues, this work proposes the \textbf{TempCompass}, a benchmark to comprehensively evaluate the temporal perception ability of Video LLMs. TempCompass introduces five basic temporal aspects (\textit{Action, Speed, Direction, Attribute Change} and \textit{Event Order}) and ten fine-grained sub-aspects, as shown in Figure \ref{fig:temporal_aspects}. Additionally, TempCompass involves four distinct types of task formats (\textit{Multi-Choice QA, Yes/No QA, Caption Matching} and \textit{Caption Generation}), as shown in Figure \ref{fig:task_types_pipeline}, which allows us to investigate how the temporal perception ability of Video LLMs varies across different task formats.

The videos in TempCompass are originated from the ShutterStock\footnote{https://www.shutterstock.com} platform. These open-domain videos cover a diverse variety of contents, ranging from human activities to natural scenarios, among others. To prevent Video LLMs from leveraging single-frame bias or language priors to complete the tasks, we construct \textit{conflicting video} pairs/triplets, within which the videos share the same static content but differ from each other in a specific temporal aspect. Given the collected videos, we derive 7,540 task instructions for the four types of tasks, using a collaboration of human annotated meta-information and LLM generation.

Due to the diverse task formats in TempCompass and the free-form nature of Video LLM responses, it is non-trivial to automatically evaluate the performance of Video LLMs. To address this challenge, we resort to the language understanding ability of LLMs for evaluation. For each type of task, we use tailored evaluation prompts for ChatGPT  (\texttt{gpt3.5-turbo}) to assess whether the Video LLM response is correct. To balance the cost and accuracy of evaluation, we also adopt some rule-based assessment methods, which are implemented prior to utilizing ChatGPT.

Based on our TempCompass benchmark, we evaluate 11 SOTA multi-modal LLMs (MLLMs), including 8 Video LLMs and 3 Image LLMs. The evaluation results reveal that the Video LLMs demonstrate a deficiency in temporal perception skills, failing to surpass their Image LLMs counterparts. We also find that the temporal perception ability of MLLMs indeed varies a lot across different task formats, which emphasizes the need to incorporate diverse task formats in the assessment process.

The main contributions of this work are summarized as follows: (1) We present a benchmark with diverse temporal aspects and task formats to comprehensively evaluate the temporal perception ability of Video LLMs. (2) We introduce conflicting videos that prevent Video LLMs from exploiting singe-frame bias or language priors. (3) We combine rule-based and LLM-based methods to efficiently and accurately evaluate the responses from Video LLMs. (4) Our empirical results reveal the weak temporal perception ability of SOTA Video LLMs.

\begin{table}[t]
 
    \centering
    \resizebox{\linewidth}{!}{
    \begin{tabular}{lccc}
         \toprule
         \multirow{2}{*}{Benchmark} & \multirow{2}{*}{\makecell{Temporal\\Diversity}} & \multirow{2}{*}{\makecell{Task\\Diversity}} & \multirow{2}{*}{\makecell{Open\\Domain}} \\ \\
         \midrule
         \textcolor{lightgray}{Video Understanding Benchmarks} \\
         MSVD-QA \citep{MSRVTT_MSVD_QA} & \crossmark & \crossmark & \checkmark  \\
         MSRVTT-QA \citep{MSRVTT_MSVD_QA} & \crossmark & \crossmark & \checkmark  \\
         TGIF-QA \citep{TGIF-QA} & \crossmark & \crossmark & \checkmark  \\
         SSv2 \citep{ssv2} & \crossmark & \crossmark & \crossmark  \\
         SSv2-label \citep{Lei2022RevealingSF} & \crossmark & \crossmark & \crossmark  \\
         CLEVRER \citep{CLEVRER} & \crossmark & \crossmark & \crossmark  \\
         ActivityNet-QA \citep{ActivityNet-qa} & \crossmark & \crossmark & \crossmark  \\
         NEXT-QA \citep{NExT-QA} & \crossmark & \checkmark & \crossmark  \\
         ViLMA \citep{ViLMA} & \checkmark & \crossmark & \checkmark  \\
         Perception Test \citep{PerceptionTest} & \checkmark & \checkmark & \crossmark  \\
         VITATECS \citep{VITATECS} & \checkmark & \crossmark & \checkmark  \\
         \midrule
         \textcolor{lightgray}{Video LLM Benchmarks} \\
         SEEDBench \citep{SEED-Bench} & \crossmark & \crossmark & \crossmark  \\
         Video-Bench \citep{Video-Bench} & \crossmark & \crossmark & \checkmark  \\
         VLM-Eval \citep{VLM-Eval} & \crossmark & \checkmark & \checkmark  \\
         AutoEval-Video \citep{AutoEval-Video} & \checkmark & \crossmark & \checkmark  \\
         MVBench \citep{MVBench} & \checkmark & \crossmark & \checkmark  \\
         \midrule
         TempCompass (Ours) & \checkmark & \checkmark  & \checkmark \\
         \bottomrule
    \end{tabular}
    }
   \caption{Comparison with related benchmarks. The rightmost three columns represent, respectively, whether the benchmark assesses performance across diverse temporal aspects, task formats, and includes open-domain videos. The detailed temporal aspects and task formats are described in Appendix \ref{app:benchmark_compare}.}
    \label{tab:benchmark_comparison}
\end{table}

\section{Related Work}
\subsection{Multi-Modal Large Language Models}
Following the success of pure-text LLMs \cite{GPT3,chatgpt,LLaMA,LLaMA-2,alpaca,Vicuna}, numerous recent efforts have been made to build multi-modal LLMs (MLLMs). To enable LLMs to comprehend visual context, two categories of paradigms have emerged and evolved. The \textit{Pipeline} paradigm \citep{HuggingGPT,ViperGPT,VisualChatGPT,MM-REACT} leverages off-the-shelf vision expert models to extract visual information in the form of texts, which are then fed to LLMs to perform the downstream vision tasks. The \textit{End-to-End} paradigm integrates vision encoders and LLM in an end-to-end trainable manner. The outputs from vision encoders are mapped to the LLM embedding space, using linear projectors \citep{LLaVA,LLaVA-v1.5,MiniGPT-4}, attention-based projections \citep{BLIP-2,mPLUG-Owl,InstructBLIP,QWen-VL-Chat} or mixed projections \citep{SPHINX,SPHINX-X}. Recent Video LLMs \citep{PandaGPT,VideoChat,Video-LLaMA,Video-LLaVA,MVBench,Video-ChatGPT,Valley,LLaMA-VID,Chat-UniVi} primarily follow the \textit{End-to-End} paradigm, with optional temporal modules to model the temporal information across frames.

\subsection{Temporal Perception Evaluation}
Temporal perception is a fundamental distinction between video-centered and image-centered applications. Prior to the age of LLMs, a lot of studies \citep{ssv2,CLEVRER,ActivityNet-qa,TestOfTime,RevisitVideo,LocalizingMoments,TemporalDataset,TGIF-QA, Ren2023TESTA,NExT-QA} have been conducted to evaluate the temporal perception performance of video-language models. However, most of these works neglect the distinction between various temporal aspects. To tackle this issue, the Perception Test \citep{PerceptionTest}, VITATECS \citep{VITATECS} and ViLMA \citep{ViLMA} introduce a diversity of fine-grained temporal aspects, thereby enabling a more comprehensive and nuanced evaluation of the temporal perception capability. However, VITATECS and ViLMA are limited in the diversity of task formats and Perception Test is constrained to indoor videos, making them less ideal to evaluate Video LLMs.

\subsection{MLLM Benchmarks}
With the advent of MLLMs, there is an increasing number of MLLMs benchmarks. A majority of them \citep{MME,MMBench,MM-Vet,TouchStone,LVLM-eHub} are specifically designed for Image LLMs. Recently, some tailored benchmarks have also been proposed for Video LLMs. However, among these Video LLM benchmarks, SEEDBench \citep{SEED-Bench}, VLM-Eval \citep{VLM-Eval} and Video-Bench \citep{Video-Bench} fall short in discerning between various temporal aspects. AutoEval-Video \citep{AutoEval-Video} and MVBench \citep{MVBench} define and incorporate a range of temporal aspects while lacking diverse task formats. 

Table \ref{tab:benchmark_comparison} compares TempCompass with representative video understanding and Video LLM benchmarks. We can see that TempCompass stands out by emphasizing diverse temporal aspects, task formats and open-domain videos.

\begin{figure}[t]
\centering
\includegraphics[width=1.\linewidth]{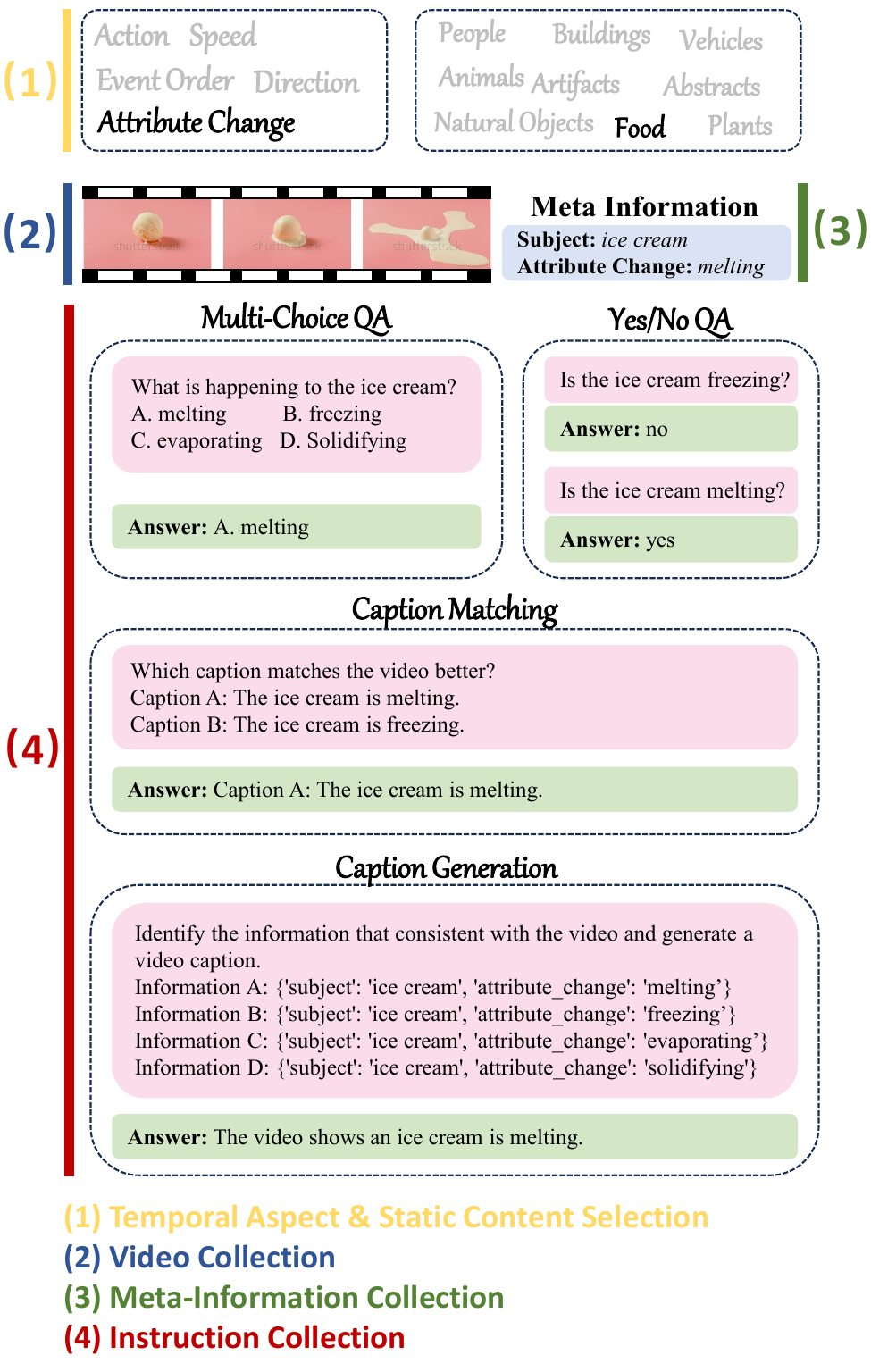}
\caption{Illustration of the four types of task formats and the data collection steps.}
\label{fig:task_types_pipeline}
\end{figure}
\begin{figure*}[t]
\centering
\includegraphics[width=1.\textwidth]{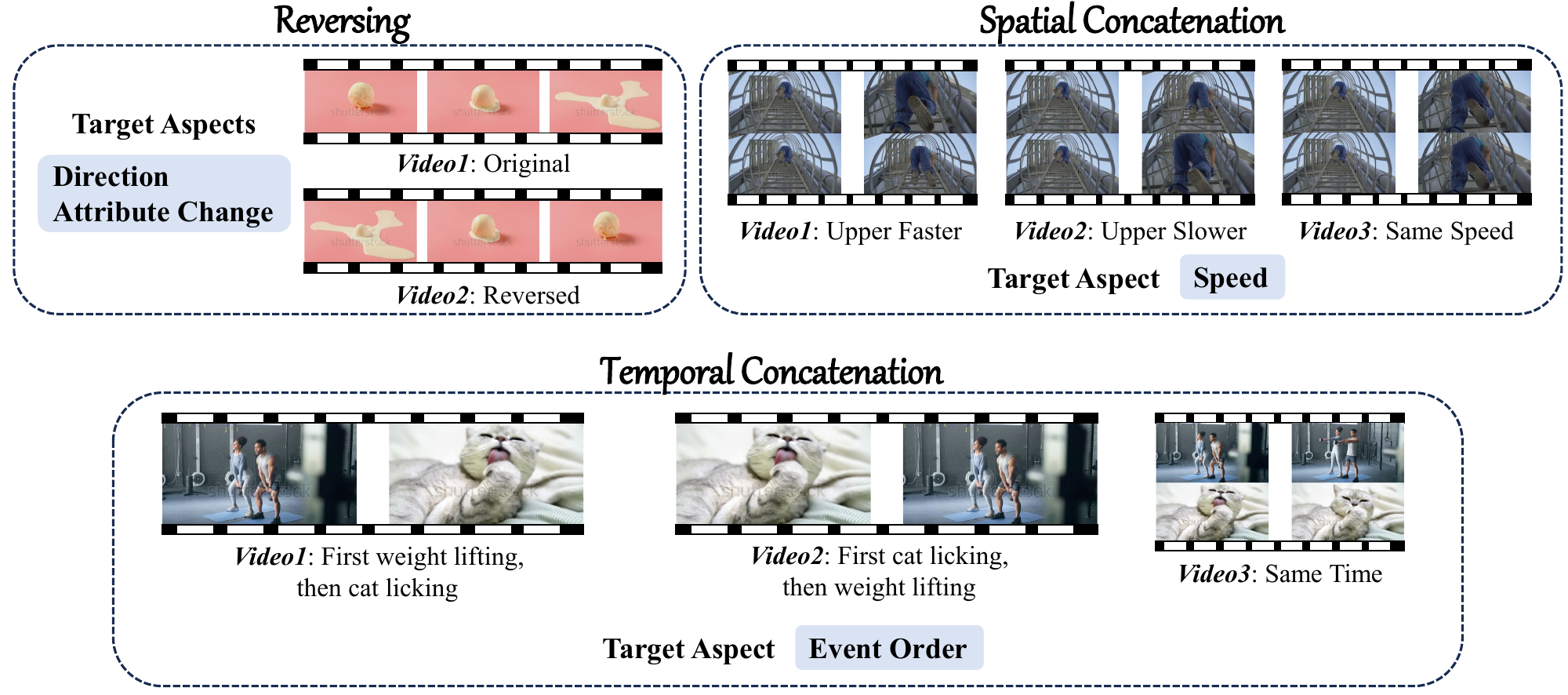}
\caption{Illustration of conflicting video pairs/triplets for different temporal aspects.}
\label{fig:conflicting_videos}
\end{figure*}

\section{TempCompass Benchmark}
TempCompass is a dataset of videos and task instructions intended to test the temporal perception ability of Video LLMs. This section will introduce the temporal aspects, task formats and static contents included in TempCompass (Section \ref{sec:benchmark_structure}), how to collect the videos and task instructions (Section \ref{sec:data_collect}) and how to automatically evaluate Video LLMs on TempCompass (Section \ref{sec:auto_eval}).

\subsection{Benchmark Structure}
\label{sec:benchmark_structure}
\subsubsection{Temporal Aspects}
\label{sec:temporal_aspects}
In contrast to images that only contain static visual information, videos convey dynamic visual information over time, i.e., temporal information. As shown in Figure \ref{fig:temporal_aspects}, we identify five basic aspects of temporal information in TempCompass:

\paragraph{Action.} This aspect assesses the ability to distinguish between different types of actions, which is a common task for video understanding models. We further divide this aspect into \textit{Coarse-Grained Action} and \textit{Fine-Grained Action}. The former involves a broader set of activities or movements while the latter is about more specific and detailed actions.

\paragraph{Speed.} This aspect delves into the capacity to discern variations in speed, which is further categorized into two components. \textit{Absolute Speed} focuses on the speed of a specific object or the pace of an entire video while \textit{Relative Speed} compares the speed of different objects. 

\paragraph{Direction.} This aspect emphasizes the perception of movement direction. Under this aspect, we separately consider the direction of objects (\textit{Object Direction}) and the direction of camera (\textit{Camera Direction}).

\paragraph{Attribute Change.} This aspect centers on how the attribute of objects or the entire video change over time. Attribute change encompasses four sub-aspects, including \textit{Size \& Shape}, \textit{Color \& Light Change}, \textit{Combined Change} and \textit{Other Change}.

\paragraph{Event Order.} This aspect focuses on the chronological order that different events happen in a video. 

\subsubsection{Task Formats}
Having established the definition of different aspects of temporal information, we now deal with the question of ``how to examine whether a Video LLM understands a specific temporal information?'' As illustrated in Figure \ref{fig:task_types_pipeline}, for a specific temporal information in the given video, we test the temporal perception ability of Video LLMs using four types of tasks: \textbf{(1)} \textit{Multi-Choice QA} asks the model to select the correct answer from multiple candidate choices. \textbf{(2)} \textit{Yes/No QA} involves the model determining whether a statement is correct based on the video. \textbf{(3)} \textit{Caption Matching} requires the model to distinguish between two video captions, one of which is consistent with the video while the other is inconsistent with the video in the temporal aspect of interest. \textbf{(4)} In the task of \textit{Caption Generation}, several pieces of information about the given temporal aspect are presented to the model, which is then asked to select the correct one and generate a video caption accordingly. Such a constrained form of captioning makes it easier to automatically evaluate the correctness of the generated caption (see Section \ref{sec:auto_eval} for details).

\subsubsection{Static Contents}
 We define nine categories of static contents: \textit{people, animals, plants, food, natural objects, vehicles, artifacts, buildings, abstract} (please refer to Appendix \ref{app:static_content} for detailed descriptions). Each video in TempCompass is classified into one or multiple categories, depending on the static visual content.

\subsection{Data Collection}
\label{sec:data_collect}
Each data example in TempCompass contains four components: video, meta-information, static content categories and task instructions. As shown in Figure \ref{fig:task_types_pipeline}, we collect these components in four steps. \textbf{(1)} We first select a set of temporal aspects and static content categories, based on which we then \textbf{(2)} collect a video together with \textbf{(3)} annotated meta-information. \textbf{(4)} Following this, we employ ChatGPT (\texttt{gpt3.5-turbo})~\citep{chatgpt}, an LLM, to generate task instructions according to the meta-information. Next, we will describe how to collect the three components in detail.

\subsubsection{Video Collection}
\label{sec:video_collect}
\begin{figure*}[t]
  \centering
  \subfigure[Temporal Aspects.]{
    \label{fig:statistic_video_temporal}
    \includegraphics[width=.49\textwidth]{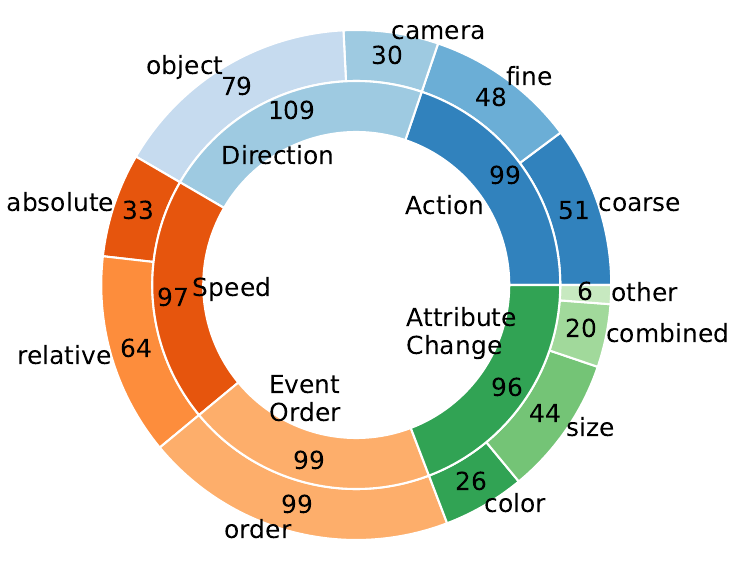}}
  \subfigure[Static Content Categories.]{
    \label{fig:statistic_video_static_content}
    \includegraphics[width=.49\textwidth]{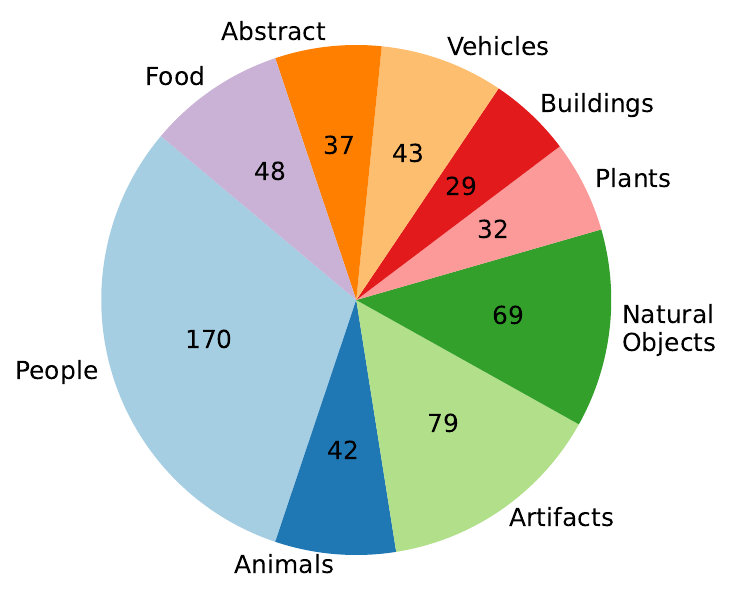}}
  \caption{Distribution of videos over temporal aspects and static content categories.}
  \label{fig:statistic_video}
\end{figure*}

We collect raw videos from the \href{https://www.shutterstock.com}{ShutterStock} platform. To enhance video diversity, we carefully control the static content distribution, guaranteeing that each category contains an adequate number of video samples. (Figure \ref{fig:statistic_video_static_content} shows the distribution). At the same time, we ensure that the videos are not included in WebVid~\citep{Bain2021FrozenIT}, a dataset widely used in pre-training video-language models. 

In the literature, it has been shown that video understanding models may utilize language priors or single-frame bias as shortcuts to obtain the correct answer, without truly understanding the temporal content of a video~\citep{Huang2018WhatMA,RevisitVideo,TemporalDataset,Lei2022RevealingSF,Girdhar2019CATERAD}. Language priors is the prior knowledge learned from language modeling (e.g., an ice cream is more likely to be melting instead of freezing). Single-frame bias refers to the reliance on static visual cues in a single frame, which strongly correlates with the correct answer (e.g., inferring the moving direction of a vehicle from its orientation in a single frame).

To mitigate the impact of such shortcuts, we construct conflicting video pairs/triplets. Within a pair/triplet, the videos have the same static content, but differ from each other in a particular temporal aspect. In this manner, the very shortcut that induces a correct answer for one video will inversely lead to an incorrect answer when applied to the conflicting counterpart. Specifically, as depicted in Figure \ref{fig:conflicting_videos}, we propose three methods to construct the conflicting videos:

\paragraph{Reversing.} Information of \textit{Direction} and \textit{Attribute Change} in a video can usually be modified by playing the video in reverse. Therefore, the conflicting video pairs for these two temporal aspects consist of an original video and its reversed counterpart.

\paragraph{Spatial Concatenation.} For the \textit{Speed} aspect, we first accelerate or decelerate a video. Then, we concatenate this modified video with the original one along the spatial dimension by (1) placing the faster version above or (2) placing the slower version above, creating two conflicting videos. We also construct a third video by concatenating two exactly same videos in the spatial dimension.

\paragraph{Temporal Concatenation.} For the \textit{Event Order} aspect, we concatenate two videos along the temporal dimension. Two conflicting videos are produced by reversing the order of the two original videos, creating two different sequences of events. Additionally, we construct a third video by spatially concatenating the two original videos, thereby presenting the two events at the same time.

\subsubsection{Meta-Information Collection}
\label{sec:meta_collect}
Given a collected video, we convert its key information into textual format. To reduce the load of annotation, we manually annotate semi-structured meta-information. As Figure \ref{fig:temporal_aspects}, \ref{fig:task_types_pipeline} shows, each piece of meta-information is comprised of two parts: (1) a phrase describing the subject and (2) another phrase describing the information related to the temporal aspect of interest.

\subsubsection{Instruction Collection}
\label{sec:instrucion_collect}
With the annotated meta-information, we obtain the task instructions via a process with interleaved automatic generation and manual refinement. Specifically, we first employ ChatGPT to automatically generate \textit{Multi-Choice QA} instructions based on the meta-information. Then, these instructions are checked and rectified by humans. Subsequently, we prompt ChatGPT to generate \textit{Yes/No QA}, \textit{Caption Matching} and \textit{Caption Generation} instructions, based on the manually rectified \textit{Multi-Choice QA} instructions. These instructions are also further checked and rectified by humans. More details of instruction collection and the prompts for instruction generation are shown in Appendix \ref{app:instruction_collect}.

\subsubsection{Data Statistics}
\label{sec:data_statistics}
We collect a total of 410 videos and 500 pieces of meta-information (a video may be annotated with multiple pieces of meta-information). Figure \ref{fig:statistic_video} depicts the video statistics, revealing an even distribution across basic temporal aspects, with roughly 100 videos representing each aspect. The nine content categories are also well covered by our collected videos. These data distributions demonstrate the diversity of TempCompass in terms of both temporal aspects and static visual contents.

\begin{table*}[t]
    \centering
    \small
    \resizebox{\linewidth}{!}{
        \begin{tabular}{lccccccccccccc}
             \toprule
              & \multicolumn{2}{c}{Baseline} & \multicolumn{3}{c}{Image LLM} & \multicolumn{8}{c}{Video LLM} \\ 
              \cmidrule(lr){2-3} \cmidrule(lr){4-6} \cmidrule(lr){7-14}
              & \multirow{2}{*}{Human} & \multirow{2}{*}{Random} & LLaVA-1.5 & SPHINX-v2 & Qwen-VL-Chat & V-LLaVA & LLaMA-VID & mPLUG-Owl & PandaGPT & Valley & VideoChat2 & V-ChatGPT & V-LLaMA \\
              & && 13B & 13B & 7B & 7B & 7B & 7B & 13B & 7B & 7B & 7B & 13B \\
             \toprule
                \textbf{Multi-Choice QA} &&&&&&&&&&&&& \\
             \midrule
                Action & 100 & 28.9 & 71.3 & \textbf{89.9} & 85.8 & 70.4 & 58.6 & 66.6 & 35.5 & 47.0 & \underline{88.5} & 47.0 & 54.1 \\
                Direction & 96.7 & 27.8 & 31.6 & \textbf{37.0} & \underline{36.7} & 32.2 & 29.9 & 29.3 & 27.8 & 29.3 & 36.4 & 31.6 & 24.5 \\
                Speed & 90 & 32.1 & 36.0 & \textbf{43.2} & \underline{42.3} & 38.2 & 29.3 & 32.2 & 29.3 & 32.5 & 42.0 & 28.4 & 28.1 \\
                Event Order & 100 & 32.2 & 34.4 & 36.4 & \underline{40.7} & \textbf{41.4} & 30.5 & 34.8 & 31.8 & 18.9 & \underline{40.7} & 37.1 & 32.8 \\
                Attribute Change & 100 & 28.5 & 38.9 & \underline{45.1} & 44.8 & 39.9 & 26.0 & 35.4 & 30.9 & 29.9 & \textbf{45.5} & 30.9 & 28.5 \\
             \midrule
                Avg & 97.3 & 29.9 & 42.8 & \underline{50.9} & 50.6 & 44.7 & 35.3 & 40.0 & 31.1 & 31.8 & \textbf{51.1} & 35.2 & 33.9 \\
                Match Rate & - & - & 84.2 & \underline{99.6} & 46.8 & 37.9 & 62.9 & 3.1 & 6.4 & 3.5 & \textbf{100.0} & 1.3 & 0.6 \\
             \bottomrule
             \toprule
                \textbf{Yes/No QA} &&&&&&&&&&&&& \\
             \midrule
                Action & 96.7 & 50.0 & 74.7 & \underline{79.1} & \textbf{81.4} & 74.3 & 63.0 & 64.4 & 53.0 & 58.1 & 72.8 & 52.5 & 68.1 \\
                Direction & 83.3 & 50.0 & 48.8 & 51.2 & 51.6 & 51.8 & 48.8 & 50.6 & 49.6 & \underline{52.0} & \textbf{53.8} & 50.0 & 46.0 \\
                Speed & 96.7 & 50.0 & 49.0 & \underline{54.7} & \textbf{59.8} & 50.3 & 49.2 & 51.2 & 50.8 & 52.5 & 53.8 & 49.5 & 48.8 \\
                Event Order & 93.3 & 50.0 & 49.5 & \textbf{54.5} & 50.8 & 49.2 & 48.4 & 51.3 & \underline{53.7} & 50.3 & 51.3 & 51.0 & 51.8 \\
                Attribute Change & 100 & 50.0 & \textbf{55.4} & 50.4 & 49.1 & 51.1 & 52.7 & 52.0 & 52.2 & 52.9 & \underline{53.8} & 50.0 & 50.9 \\
                \midrule
                Avg & 94 & 50.0 & 56.4 & \underline{59.1} & \textbf{60.0} & 56.4 & 53.0 & 54.4 & 51.8 & 53.5 & 58.0 & 50.7 & 53.7 \\
                Match Rate & - & - & \textbf{100.0} & \textbf{100.0} & \underline{99.8} & \textbf{100.0} & 99.1 & 95.6 & \textbf{100.0} & 98.7 & 18.8 & \textbf{100.0} & 95.1 \\
             \bottomrule
             \toprule
                \textbf{Caption Matching} &&&&&&&&&&&&& \\
             \midrule
                Action & 100 & 50.0 & 86.9 & \underline{89.2} & \textbf{90.2} & 88.2 & 72.7 & 56.9 & 56.6 & 15.5 & 65.0 & 64.6 & 73.1 \\
                Direction & 96.7 & 50.0 & 50.8 & 52.0 & \underline{53.5} & \textbf{53.8} & 45.6 & 45.3 & 51.4 & 21.4 & \textbf{53.8} & 48.6 & 47.4 \\
                Speed & 100 & 50.0 & 54.6 & 47.1 & \underline{55.0} & \textbf{61.9} & 52.2 & 46.4 & 44.3 & 22.0 & 52.6 & 47.8 & 47.1 \\
                Event Order & 100 & 50.0 & 55.0 & 53.0 & \textbf{60.3} & \underline{57.0} & 49.0 & 49.3 & 55.0 & 28.3 & 53.0 & 49.3 & 52.0 \\
                Attribute Change & 100 & 50.0 & 51.0 & 55.2 & \underline{56.9} & \textbf{58.3} & 49.0 & 49.0 & 49.0 & 22.9 & 53.8 & 48.6 & 48.3 \\
                \midrule
                Avg & 99.3 & 50.0 & 59.5 & 59.2 & \underline{63.1} & \textbf{63.7} & 53.6 & 49.3 & 51.3 & 22.0 & 55.6 & 51.8 & 53.5 \\
                Match Rate & - & - & 91.2 & 89.3 & \underline{91.6} & 76.6 & 44.5 & 15.8 & 30.7 & 11.2 & \textbf{95.3} & 7.5 & 0.1 \\
             \bottomrule
             \toprule
                \textbf{Caption Generation} &&&&&&&&&&&&& \\
                \midrule
                Action & 100 & 28.8 & \underline{67.4} & \textbf{67.9} & 62.6 & 50.8 & 53.0 & 46.5 & 23.7 & 24.7 & 54.0 & 40.9 & 54.3 \\
                Direction & 86.7 & 28.4 & \textbf{31.9} & 19.0 & 27.8 & 28.7 & 28.0 & 28.2 & 25.7 & 20.4 & \underline{31.0} & 28.4 & 21.3 \\
                Speed & 100 & 32.4 & 24.7 & 20.4 & 29.6 & 23.2 & 21.9 & \underline{30.4} & 26.0 & 21.9 & \textbf{32.7} & 24.5 & 13.9 \\
                Event Order & 100 & 32.1 & 33.0 & 37.2 & 34.8 & \underline{38.2} & 35.5 & 31.2 & 29.8 & 35.8 & 34.2 & 31.8 & \textbf{38.5} \\
                Attribute Change & 100 & 28.6 & 35.4 & 31.0 & 32.3 & 33.6 & 35.9 & \underline{36.5} & 32.6 & 29.4 & \textbf{41.4} & 33.9 & 33.9 \\
                \midrule
                Avg & 97.3 & 30.0 & \underline{38.4} & 34.9 & 37.3 & 34.8 & 34.8 & 34.4 & 27.5 & 26.3 & \textbf{38.5} & 31.8 & 32.2 \\
             \bottomrule
        \end{tabular}
    }
    \caption{Accuracy of MLLMs on our TempCompass benchmark. ``V-'' in the model names stands for ``Video-''. The best and second-best MLLM results are \textbf{bold} and \underline{underlined}, respectively. "Match Rate" denotes the success rate of matching a predicted option from the MLLM's response using hand-crafted rules. The complete results of all temporal aspects are reported in Appendix \ref{app:quantative_results}.}
    \label{tab:eval_results_coarse_compress}
\end{table*}

Given a piece of meta-information, we collect multiple instructions for each type of task: at least 3 for \textit{Multi-Choice QA}, 2 for \textit{Yes/No QA}, 3 for \textit{Caption Matching}, and 4 for \textit{Caption Generation}. In this way, we collect a total of 7,540 instructions in our benchmark. In Appendix \ref{app:statistic}, we show the detailed distribution of task instructions, video duration and answer distribution. In Appendix \ref{app:data_exmaple}, we present complete data examples including the video, meta-information, static content and instructions.

\subsection{Quality Verification}
After the data collection process described in Section \ref{sec:data_collect}, we randomly sample 200 task instructions to verify the data quality. These instructions and videos are presented to three human annotators to perform the task. Human annotators also have the option to label an instruction as "Cannot Answer", which indicates that the instruction is unreasonable. Among the 600 annotated results, only 5 are labeled as "Cannot Answer". Table \ref{tab:eval_results_coarse_compress} also show that the human annotators achieve near-perfect accuracy across most tasks and aspects, attesting to the high quality of the collected data. More details of quality verification can be found in Appendix \ref{app:quality_verify_human_baseline}.

\subsection{Automatic Evaluation}
\label{sec:auto_eval}
For \textit{Multi-Choice QA}, \textit{Yes/No QA} and \textit{Caption Matching}, we adopt a hybrid approach that integrates rule-based methods and ChatGPT to automatically evaluate the responses generated by Video LLMs. To begin with, we check whether any candidate option (e.g., \textit{A/B/C/D}, \textit{Yes/No} or \textit{Caption A/Caption B}) is explicitly mentioned in the response and compare it against the ground-truth answer. Hand-crafted matching rules are specifically designed for different types of tasks. Then, for responses that fail to match any candidate options, we resort to ChatGPT's language understanding ability to determine whether they are correct based on the task instruction and ground-truth answer. Details of the matching rules and the prompts for LLM-based evaluation are illustrated in Appendix\ref{app:auto_eval}.

When it comes to the \textit{Caption Generation} task, the rule-based evaluation method is ineffective because almost all Video LLM responses are free-form video captions. Therefore, we solely rely on ChatGPT for evaluation. Specifically, we prompt ChatGPT to answer the corresponding \textit{Multi-Choice} question using the generated video caption as context. If the answer by ChatGPT is correct, then the generated caption is deemed as correct and vice versa. The motivation is that if the Video LLM selects an incorrect information to generate the caption, ChatGPT will consequently select an incorrect option. Considering the possibility that the generated caption may not involve any of the provided information, we include an extra option: ``\textit{None of the choices are correct}'' in the \textit{Multi-Choice} question. In case where ChatGPT selects this option, the generated caption is also deemed as incorrect.

\section{Experiments}
\subsection{Evaluated Models}
We conduct evaluation experiments on a total of 11 open-sourced state-of-the-art MLLMs, including Video-LLaMA \citep{Video-LLaMA}, Video-ChatGPT \citep{Video-ChatGPT}, Valley \citep{Valley}, VideoChat2 \citep{MVBench}, mPLUG-Owl \citep{mPLUG-Owl}, PandaGPT \citep{PandaGPT}, Video-LLaVA \citep{Video-LLaVA}, LLaMA-VID \citep{LLaMA-VID}, LLaVA-v1.5 \citep{LLaVA-v1.5}, SPHINX \citep{SPHINX,SPHINX-X} and Qwen-VL-Chat \citep{QWen-VL-Chat}. These models cover a wide range of Video LLMs and Image LLMs with different model architectures and training strategies. Inspired by \cite{MVBench}, we append answer prompts (e.g., ``\textit{Best Option:}'') to the task instructions to guide MLLMs generating responses in the desired formats (see Appendix \ref{app:infer_setting} for details). In addition to the MLLMs, we also incorporate random and human baselines. Details of the models and human baseline are described in Appendix \ref{app:model_detail} and Appendix \ref{app:quality_verify_human_baseline}, respectively.

\subsection{Main Results}
Table \ref{tab:eval_results_coarse_compress} summarizes the results across the four tasks. We discuss the results from four perspectives:

\paragraph{Overall Performance.} Existing MLLMs exhibit poor temporal perception ability. Five Video LLMs, i.e., LLaMA-VID, Panda-GPT, Valley, Video-ChatGPT, and Video-LLaMA, fail to convincingly surpass the random baseline across all tasks. Although Video-LLaVA and VideoChat2 exhibit improved performance, they still fall significantly short of the human. Notably, all Video LLMs struggle to consistently surpass SPHINX-v2 and Qwen-VL-Chat, two Image LLMs, highlighting a pervasive lack of temporal perception ability in current Video LLMs. This finding echoes with VITATECS \cite{VITATECS}, which reveals that current video-language models barely surpass random guesses in a task similar to our \textit{Caption Matching}.

\paragraph{Performance Across Temporal Aspects.} MLLMs demonstrate their highest proficiency in \textit{Action} aspect, with the best model achieving near 90 accuracy on \textit{Multi-Choice QA} and \textit{Caption Matching}. The reason is that the type of action can largely be deduced from static visual cues alone. This observation indicates that existing MLLMs already demonstrate a strong understanding capability of static visual information, which is the foundation to develop temporal perception capabilities. In comparison, the performance are significantly worse on the remaining four aspects, as they are more dependent on the temporal information across frames. This finding implies that there is a pressing need for enhancing the current MLLMs' capabilities in perceiving \textit{Speed}, \textit{Direction}, \textit{Event Order} and \textit{Attribute Change}.

\paragraph{Performance Across Tasks.} Comparing the results across all four tasks, we can see that there exists a significant variation in performance. This variation can be attributed to two factors. On the one hand, the inherent complexity of the tasks varies, as exemplified by the performance differences between \textit{Multi-Choice QA} and \textit{Caption Generation}.  The latter generally yields worse results, because it necessitates not only selecting the correct information but also generating the caption accordingly. On the other hand, individual models have innate strengths and weaknesses in different tasks. For instance, Video-LLaVA takes a leading place in the \textit{Speed} aspect on \textit{Caption Matching}, while performing not better than random in the same temporal aspect on \textit{Yes/No QA} and \textit{Caption Generation}. These findings suggest that the temporal perception ability displayed by MLLMs is highly dependent on the form of evaluation tasks, which emphasizes the need to incorporate a diverse array of tasks in the assessment process.

\paragraph{Ability to Respond in Desired Format.} Despite the use of answer prompts, some MLLM usually fail to respond in the desired format, as reflected by the low match rate in Table \ref{tab:eval_results_coarse_compress}. This phenomenon demonstrates the limitation of rule-based matching in evaluating MLLM responses and underlines the necessity of LLM-based evaluation. We also observe that the design of answer prompt has a non-negligible impact on the match rate. Please refer to Appendix \ref{app:answer_prompt} for the analytical study.

\begin{table}[t]
    \centering
    \begin{tcolorbox}  
        {\color[HTML]{EF0000} Question:} What is happening in the video?\\A. A person drops down the pineapple\\B. A person pushes forward the pineapple\\C. A person rotates the pineapple\\D. A person picks up the pineapple \\
        
        \textbf{Original Video}\\
        \includegraphics[width=1.\textwidth]{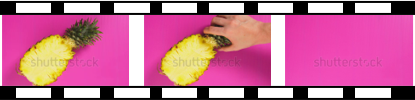} \\
        {\color[HTML]{00CC00} Ground-Truth:} D. A person picks up the pineapple\\
        {\color[HTML]{0000EF} SPHINX-v2:} D (\checkmark)\\
        {\color[HTML]{0000EF} Video-LLaVA:} D (\checkmark)\\
        {\color[HTML]{0000EF} Video-ChatGPT:} The best option is to pick up the pineapple. (\checkmark)\\
        
        \textbf{Reversed Video}\\
        \includegraphics[width=1.\textwidth]{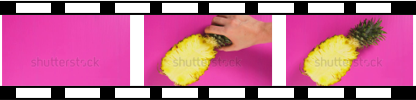} \\  
        {\color[HTML]{00CC00} Ground-Truth:} A. A person drops down the pineapple\\
        {\color[HTML]{0000EF} SPHINX-v2:} D (\crossmark)\\
        {\color[HTML]{0000EF} Video-LLaVA:} C (\crossmark)\\
        {\color[HTML]{0000EF} Video-ChatGPT:} The best option would be to pick up the pineapple and place it on the table. (\crossmark)\\
    \end{tcolorbox}
    \caption{Example of MLLM responses to a multi-choice question, given a pair of conflicting videos. \checkmark and \crossmark are assessed by our automatic evaluation method.}
    \label{tab:eval_example_simple}
\end{table}

\subsection{Qualitative Results}
Table \ref{tab:eval_example_simple} illustrates the responses from three MLLMs, given a pair of conflicting videos of the \textit{Direction} aspect. We can see that all three MLLMs accurately respond to the question when presented with the original video; however, they fail to deliver correct answers when confronted with the reversed version. This result indicates the inherent inability of the models to perceive and understand the direction of movement. The automatic evaluation results also showcase that our LLM-based evaluation method is able to deal with the free-form response from MLLMs. More qualitative results on other temporal aspects and task formats can be found in Appendix \ref{app:more_qualitative_results}.

\subsection{Automatic Evaluation Accuracy}
To validate the reliability of the proposed automatic evaluation method, we compare its results with human evaluation. The details of evaluation setups is described in Appendix \ref{app:human_eval}. Table \ref{tab:auto_eval_acc} shows the percentage of automatic evaluation results that agree with human judgements, averaged over three human evaluators. We can see that our automatic evaluation method achieves very high consistency with humans in \textit{Multi-Choice QA}, \textit{Yes/No QA} and \textit{Caption Matching}. In terms of \textit{Caption Generation}, roughly 20\% of the LLM-based evaluation are inconsistent with humans. This is because the MLLMs may hallucinate some contents irrelevant to the video, which is hard to detect for the pure-text GPT3.5-Turbo. In Appendix \ref{app:more_qualitative_results}, we present qualitative examples to better illustrate the pros and cons of our automatic evaluation method.

\begin{table}[t]
    \centering
    \resizebox{0.8\linewidth}{!}{
    \begin{tabular}{lccc}
         \toprule
         \multirow{2}{*}{Multi-Choice} & \multirow{2}{*}{\makecell{Yes/No}} & \multirow{2}{*}{\makecell{Caption\\Matching}} & \multirow{2}{*}{\makecell{Caption\\Generation}} \\ \\
         \midrule
         99.67 & 98.33 & 99.0 & 79.33  \\
         \bottomrule
    \end{tabular}
    }
   \caption{Accuracy of the automatic evaluation results, benchmarked against human evaluation as ground-truth.}
    \label{tab:auto_eval_acc}
\end{table}
\begin{table}[t]
    \centering
    \resizebox{\linewidth}{!}{
    \begin{tabular}{lcccc}
         \toprule
         \multirow{2}{*}{Model} & \multirow{2}{*}{Multi-Choice} & \multirow{2}{*}{\makecell{Yes/No}} & \multirow{2}{*}{\makecell{Caption\\Matching}} & \multirow{2}{*}{\makecell{Caption\\Generation}} \\ \\
         \midrule
         \textcolor{lightgray}{LLaVA-1.5} \\
         w/ Conflicting       & 35.1 & 50.6 & 52.8 & 31.3  \\
         w/o Conflicting    & \textbf{41.1}  & \textbf{52.8}  & \textbf{57.3}  & \textbf{33.6} \\
         \midrule
         \textcolor{lightgray}{SPHINX-v2} \\
         w/ Conflicting       & 40.3 & 52.7 & 51.8 & 26.7 \\
         w/o Conflicting    & \textbf{52.8} & \textbf{58.3} & \textbf{62.8} & \textbf{31.5}  \\
         \midrule
         \textcolor{lightgray}{Qwen-VL-Chat} \\
         w/ Conflicting   & 41.0 & 53.2 & 56.4  & 31.0 \\
         w/o Conflicting   & \textbf{49.0} & \textbf{59.5} & \textbf{63.5} &  \textbf{32.9}\\
         \midrule
         Random   & 30.1 & 50.0 & 50.0  & 30.3 \\
         \bottomrule
    \end{tabular}
    }
   \caption{The performance of Image LLMs w/ and w/o conflicting videos. The results are averaged over all temporal aspects except for the \textit{Action} aspect, for which we do not construct conflicting videos.}
    \label{tab:conflict_video_effect}
\end{table}

\subsection{Effect of the Conflicting Videos}
Table \ref{tab:conflict_video_effect} compares the performance of Image LLMs on all videos and when excluding the constructed conflicting videos (i.e., on the raw videos). Evidently, the Image LLMs notably outperform the random baseline on raw video samples, especially in \textit{Multi-Choice QA}, \textit{Yes/No QA} and \textit{Caption Matching}. This implies that, to a considerable degree, questions about raw videos can be answered by leveraging the single-frame bias and language priors. With the introduction of conflicting video, the performance of Image LLMs is clearly closer to random baseline, effectively alleviating the impact of biases. The effect of conflicting videos is also illustrated by the example cases in Table \ref{tab:eval_example_simple}, \ref{tab:eval_example_9}, \ref{tab:eval_example_10}, \ref{tab:eval_example_1}, \ref{tab:eval_example_2}.

\section{Conclusions}
In this work, we propose the TempCompass benchmark to evaluate the temporal perception ability of Video LLMs. Our benchmark introduces ten temporal aspects and four distinct types of task formats, which offers a comprehensive view to investigate the temporal perception capability. Two innovative strategies are devised in the data collection process, including (1) the construction of conflicting videos to mitigate the influence of single-frame bias and language-priors and (2) the collaboration of human annotation and LLM generation to efficiently collect high-quality task instructions. We also propose an automatic evaluation method based on ChatGPT, which is able to accurately assess the free-form Video LLM responses. Based on TempCompass, we extensively evaluate 8 SOTA Video LLMs and 3 Image LLMs. Our evaluation results reveal the pressing need to enhance the temporal perception ability of Video LLMs.

\section{Limitations}
Despite the contributions made by TempCompass, this work is still limited in two perspective. First, despite our effort in constructing the conflicting videos, the influence of single-frame and languages-priors persist. This is evident from the fact that Image LLMs continue to perform clearly above random baselines in specific tasks and temporal aspects. Second, our automatic evaluation method encounters challenges in accurately assessing certain generated video captions, which, although consistent with the ground-truth candidate information, incorporate elements of hallucinated content.

\section*{Acknowledgements}
We thank all the anonymous reviewers for their constructive comments. This work is supported in part by a Huawei Research Grant and National Natural Science Foundation of China (No. 62176002). Xu Sun is the corresponding author of this paper.

\bibliography{acl_latex}

\appendix

\section{More Details of Data}
\subsection{Static Contents}
\label{app:static_content}
Our benchmark covers nine categories of static contents, including \textit{people, animals, plants, food, natural objects, vehicles, artifacts, buildings, abstract}. \textit{Natural objects} denotes lifeless natural objects and scenery. \textit{Artifacts} encompasses human-made objects, excluding large-size objects like vehicles and buildings. \textit{Abstract} refers to abstract geometric shapes and symbols. For better understanding, please refer to the example videos with annotated categories in Table \ref{tab:data_example_1},\ref{tab:data_example_2},\ref{tab:data_example_3},\ref{tab:data_example_4}.

\subsection{Instruction Collection}
\label{app:instruction_collect}
We collect the task instructions in four steps:
\begin{enumerate}
    \item Generating \textit{Multi-Choice QA} instructions based on meta-information, using ChatGPT.
    \item Manually review and rectify \footnote{All human annotation and evaluation in this study were done by the authors.} the generated \textit{Multi-Choice QA} instructions.
    \item Generating instructions for the other three tasks based on manually rectified \textit{Multi-Choice QA} instructions, using ChatGPT.
    \item Manually review and rectify the generated instructions.
\end{enumerate}
The detailed collection process for each type of task is described as follows:
\paragraph{Multi-Choice QA} The task instructions are directly generated from the annotated meta-information. We also design some in-context learning examples to help ChatGPT better understand the task to accomplish. The detailed prompt is shown in Table \ref{tab:instrut_gen_multi_choice}. For each piece of meta-information, we prompt ChatGPT to generate five \textit{Multi-Choice QA} instructions. To prevent bias towards any specific option position, we randomly shuffle the order of the options. Following this step, the generated instructions undergo meticulous review and refinement by the authors, ensuring that a minimum of three high-quality instructions are retained within in the benchmark.

\paragraph{Yes/No QA.} Based on the manually rectified \textit{Multi-Choice QA} questions, we prompt ChatGPT to directly generate an equal number of Positive and Negative questions, as shown in Table \ref{tab:instrut_gen_yes_no}. 

\paragraph{Caption Matching.} Based on the manually rectified \textit{Multi-Choice QA} questions, we first prompt ChatGPT to generate a True caption and three False captions, which are subsequently integrated into several templates to construct the task instructions. To eliminate bias stemming from caption position, we randomize the sequence in which True and False captions are displayed for each instruction. The caption generation prompt and instruction templates are shown in Table \ref{tab:instrut_gen_caption_matching}.

\paragraph{Caption Generation.} As shown in Table \ref{tab:instrut_gen_captioning}, the instructions for this task consists of a task description and several pieces of information similar to the meta-information. We first manually compose a task description and paraphrase it using ChatGPT. Then, an instruction "Ensure that the generated video caption is brief" is appended to the two task descriptions, resulting in four task descriptions in total. The candidate information are derived from the meta-information and the manually rectified \textit{Multi-Choice} question.

\begin{figure*}[t]
\centering
\includegraphics[width=1.\textwidth]{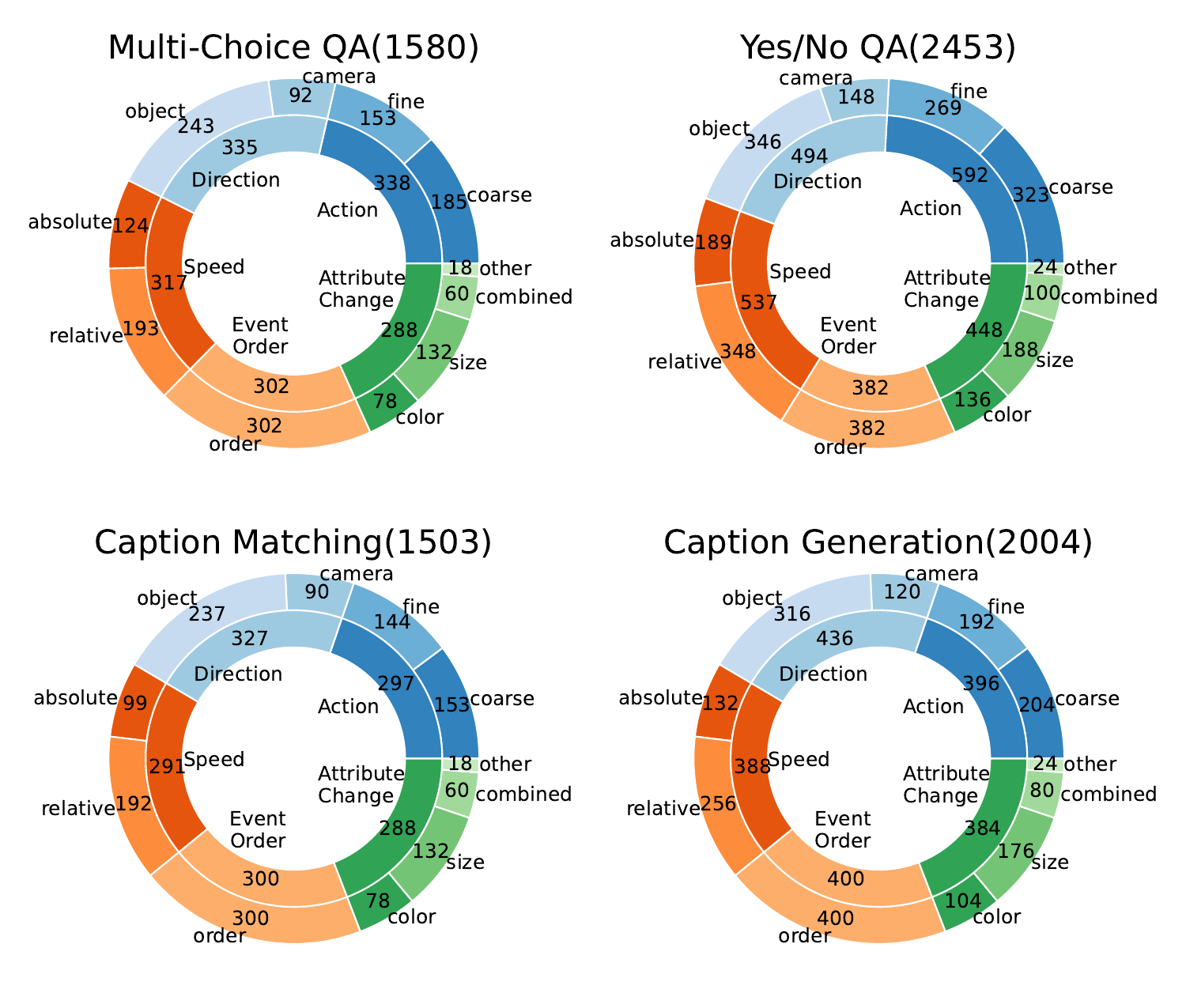}
\caption{Distribution of task instructions over the temporal aspects.}
\label{fig:statistic_question}
\end{figure*}
\begin{figure*}[t]
\centering
\includegraphics[width=1.\linewidth]{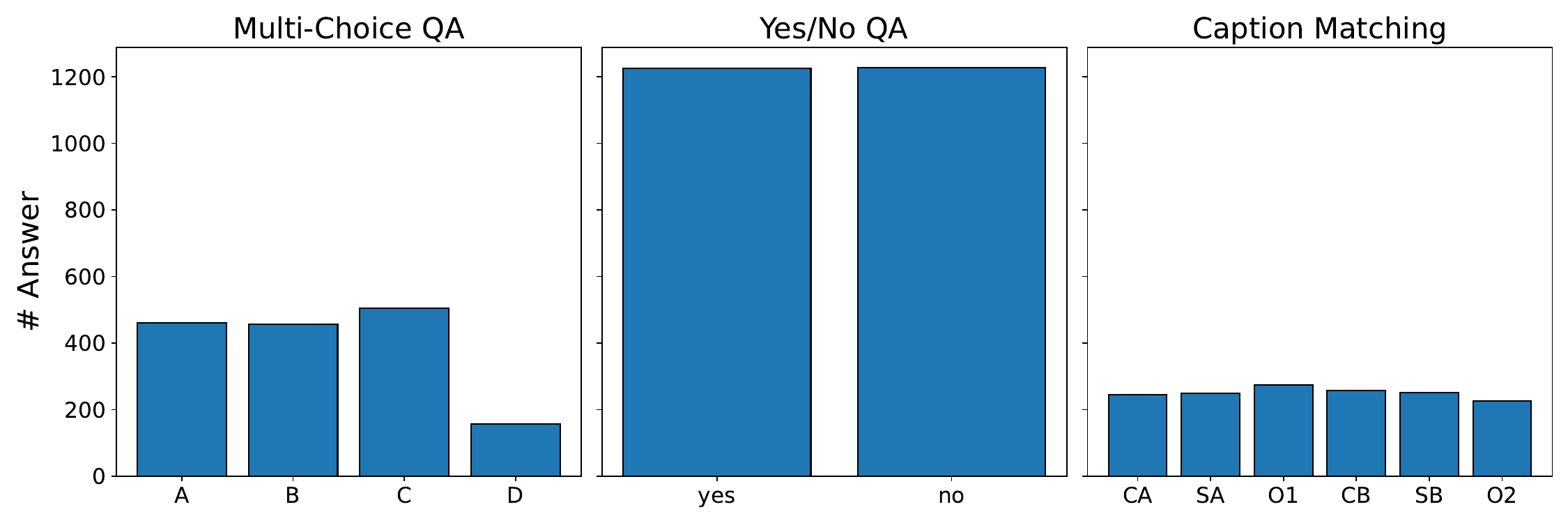}
\caption{Distribution of answers. CA, CB, SA, SB, O1, O2 stand for Caption A, Caption B, Sentence A, Sentence B, Option 1, Option 2, respectively.}
\label{fig:ans_distribution}
\end{figure*}
\begin{figure}[t]
\centering
\includegraphics[width=1.\linewidth]{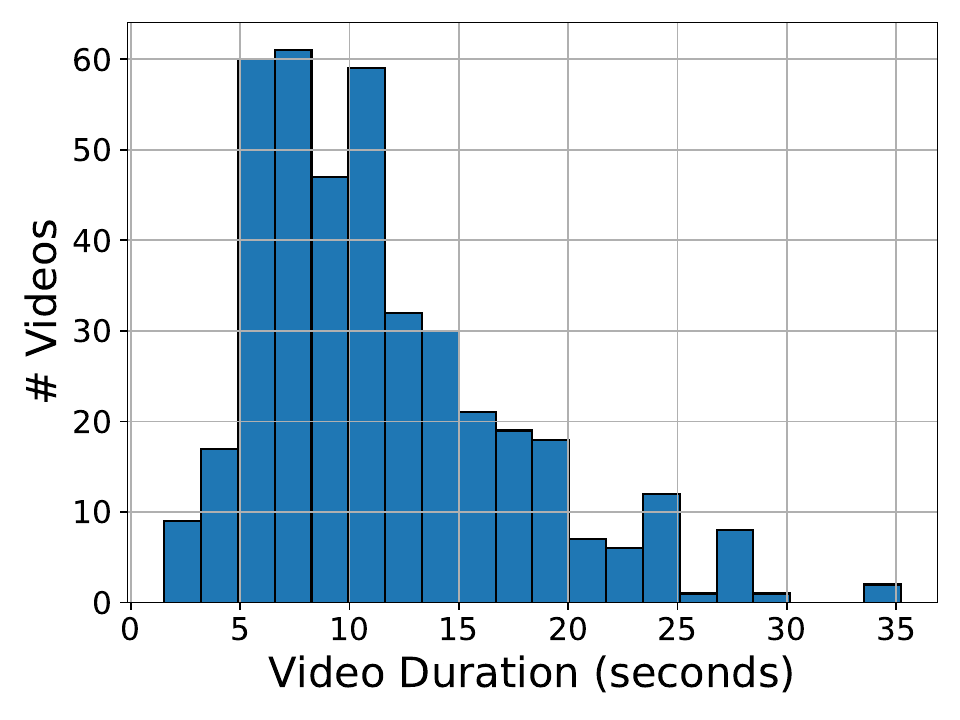}
\caption{Distribution of video duration.}
\label{fig:vid_duration}
\end{figure}
\begin{table*}[t]
    \centering
    \resizebox{\textwidth}{!}{
    \begin{tabular}{lccc}
         \toprule
         Benchmark & Temporal Aspects & Task Formats & Open Domain \\
         \midrule
         \textcolor{lightgray}{\textit{Conventional Video Understanding Benchmarks}} &&& \\
         \quad MSVD-QA \citep{MSRVTT_MSVD_QA} & - & Free-form QA & \checkmark  \\
         \quad MSRVTT-QA \citep{MSRVTT_MSVD_QA} & - & Free-form QA & \checkmark  \\
         \quad TGIF-QA \citep{TGIF-QA} & Repetition,Event Order & Free-form QA & \checkmark  \\
         \quad SSv2 \citep{ssv2} & - & \textcolor{lightgray}{Action Recognition} & \crossmark  \\
         \quad SSv2-label \citep{Lei2022RevealingSF} & - & \textcolor{lightgray}{Caption Matching} & \crossmark  \\
         \quad CLEVRER \citep{CLEVRER} & - & MC QA,Free-form QA & \crossmark  \\
         \quad ActivityNet-QA \citep{ActivityNet-qa} & Action,Event Order & Free-form QA & \crossmark  \\ \\
         \quad NEXT-QA \citep{NExT-QA} & Action,Event Order & MC QA,Free-form QA & \crossmark  \\ \\
         \quad ViLMA \citep{ViLMA} & \makecell{Action,Direction,\\X Change,Repetition} & Caption Matching & \checkmark  \\ \\
         \quad \multirow{3}{*}{\makecell{VITATECS \citep{VITATECS}}} & \multirow{3}{*}{\makecell{Action,Event Order,\\Speed,Direction\\Object Interaction}} & \multirow{3}{*}{\makecell{Caption Matching}} & \multirow{3}{*}{\makecell{\checkmark}}  \\ \\ \\ \\
         \quad \multirow{3}{*}{\makecell{Perception Test \citep{PerceptionTest}}} & \multirow{3}{*}{\makecell{Event Order,Repetition,\\Direction,Action,X Change\\Temporal Localization}} & \multirow{3}{*}{\makecell{MC QA,Grounded Video QA\\\textcolor{lightgray}{Object Tracking,Point Tracking}\\ \textcolor{lightgray}{Action Localization}}} & \multirow{3}{*}{\makecell{\crossmark}}  \\ \\ \\
         \midrule
         \textcolor{lightgray}{\textit{Video LLM Benchmarks}} &&& \\
         \quad SEEDBench \citep{SEED-Bench} & Action,Event Order & MC QA & \crossmark  \\
         \quad Video-Bench \citep{Video-Bench} & - & MC QA & \checkmark  \\ \\
         \quad \multirow{2}{*}{\makecell{VLM-Eval \citep{VLM-Eval}}} & \multirow{2}{*}{\makecell{-}} & \multirow{2}{*}{\makecell{Free-form QA,Retrieval,\\Caption Generation}} & \multirow{2}{*}{\makecell{\checkmark}}  \\ \\ \\
         \quad \multirow{2}{*}{\makecell{AutoEval-Video \citep{AutoEval-Video}}} & \multirow{2}{*}{\makecell{Event Order,Direction,\\Attribute Change}} & \multirow{2}{*}{\makecell{Free-form QA}} & \multirow{2}{*}{\makecell{\checkmark}}  \\ \\ \\
         \quad \multirow{4}{*}{\makecell{MVBench \citep{MVBench}}} & \multirow{4}{*}{\makecell{Action,Repetition,Direction,\\Temporal Localization,\\X Change,Event Order\\Object Interaction}} & \multirow{4}{*}{\makecell{MC QA}} & \multirow{4}{*}{\makecell{\checkmark}}  \\ \\ \\ \\

        \midrule
         \quad \multirow{3}{*}{TempCompass (Ours)} & \multirow{3}{*}{\makecell{Action,Speed,Direction,\\Attribute Change,Event Order}} & \multirow{3}{*}{\makecell{MC QA,Y/N QA,\\Caption Matching,\\Caption Generation}}  & \multirow{3}{*}{\makecell{\checkmark}} \\ \\ \\
         \bottomrule
    \end{tabular}
    }
   \caption{Comparison with related benchmarks. The temporal aspects focus on basic temporal perception ability while excluding the aspects that require reasoning skills. "MC QA" and "Y/N QA" represent multi-choice QA and Yes/No QA, respectively. Video LLMs cannot be directly tested on the \textcolor{lightgray}{gray} task formats because they lack textual task instructions. Detailed definition of some temporal aspects and task formats are explained in Appendix \ref{app:benchmark_compare}.}
    \label{tab:benchmark_comparison_detail}
\end{table*}

\subsection{Data Statistics}
\label{app:statistic}
\paragraph{Task Instructions.} Figure \ref{fig:statistic_question} presents the distribution of task instructions. We can see that each type of task involves at least 1,500 instructions and every basic temporal aspect has a balanced number of these instructions.

\paragraph{Answers.} As can be seen in Figure \ref{fig:ans_distribution}, the distribution of ground-truth answers within our benchmark dataset is balanced across all options. An exception is the option "\textit{D}" in \textit{Multi-Choice QA}, which appears less frequently compared to the other three options. This is because not all \textit{Multi-Choice} question includes four options. when we restrict our analysis to questions that offer exactly four options (a total of 675 instances), the frequency of "\textit{D}" as the correct answer (occurring 157 times) aligns closely with the frequencies of the remaining three options.

\paragraph{Video Duration.} Figure \ref{fig:vid_duration} shows the distribution of video duration. Our benchmark primarily focuses on short and medium-length videos within 30 seconds.

\subsection{Quality Verification and Human Baseline}
\label{app:quality_verify_human_baseline}
\begin{figure}[t]
\centering
\includegraphics[width=1.\linewidth]{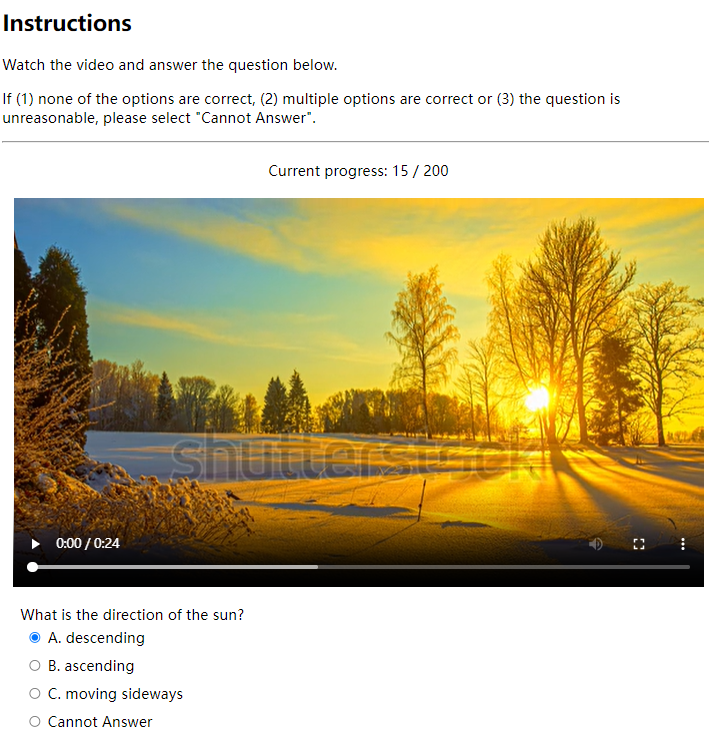}
\caption{Screenshot of the interface to collect human answers.}
\label{fig:human_perform_screenshot}
\end{figure}

We randomly sample 200 task instructions, with a balanced distribution of 10 instructions for each temporal aspect across every task (i.e., 50 instructions per task). For \textit{Multi-Choice QA}, \textit{Yes/No QA} and \textit{Caption Matching}, human annotators are directly asked to select an option, instead of generating a free-form answer as the MLLMs. The selected option is then compared with the ground-truth answer. For \textit{Caption Generation}, human annotators follow the same instructions presented to MLLMs to generate video captions, which are then evaluated in the manner described in Appendix \ref{app:auto_eval}. In addition to performing the task, human annotators have another option to label a task instruction as "Cannot Answer". In this case, the answer is considered as incorrect when evaluating the human performance. Figure \ref{fig:human_perform_screenshot} shows the interface to collect human answers. The final results are obtained by averaging among three human annotators.

\subsection{Data Examples}
\label{app:data_exmaple}
Table \ref{tab:data_example_1},\ref{tab:data_example_2},\ref{tab:data_example_3},\ref{tab:data_example_4} illustrate complete data examples in our benchmark. Each example contains the video, meta-information, static content categories and task instructions.

\subsection{Comparison With Related Benchmarks}
\label{app:benchmark_compare}
Table \ref{tab:benchmark_comparison_detail} summarizes the specific temporal aspects and task formats involved in related benchmarks. We can see that the majority of existing benchmarks lack a comprehensive categorization of temporal aspects. By contrast, VITATECS, Perception Test, ViLMA and MVBench introduce a variety of temporal aspects, which are complementary to the ones presented in our TempCompass. Meanwhile, the variation in performance across different task formats cannot be reflected by most current benchmarks. While Perception Test considers both multiple task formats and temporal aspects, it is constrained to indoor videos that focusing on people and artifacts. In comparison, our proposed TempCompass uniquely stands out by emphasizing a rich variety of temporal dimensions, diverse task formats and open-domain videos. This design enables TempCompass to provide a more holistic assessment of Video LLM's temporal perception capabilities.

It is worth noting that the definition of temporal aspects and task formats vary among different studies. For the sake of clarity, we unify their naming in Table \ref{tab:benchmark_comparison_detail}. Here we explain some definitions as follows:

\paragraph{Temporal Aspects.} "Repetition" measures the ability to count the number of repeating activities. "Object Interaction" focuses on the relationship between different objects participating in the same event. "Temporal Localization" require the model to identify the temporal position of specific events in the video. "X Change" encompasses various changes over time, including attribute change, scene change, etc.

\paragraph{Task Formats.} "Free-form QA" may involve different formats of task instructions but a proper categorization is not provided in the benchmark. In "Action Recognition", the model is required to classify videos into a predetermined set of actions. Notably, the original SSv2 dataset does not offer explicit task instructions for this classification process. "Grounded Video QA" demands that the model tracks the objects meeting specific conditions by pinpointing them within bounding boxes throughout the video. "Object Tracking" and "Point Tracking" require tracking the bounding boxes and points, without providing a textual task instruction. "Retrieval" encompasses text-to-video (T2V) retrieval and video-to-text (V2T) retrieval. Taking V2T as example, the Video LLM first generates a description of the video, which is then used to retrieve the relevant texts.

\section{More Details of Evaluation Setups}
\label{app:auto_eval}
\subsection{Rule-based Evaluation}
For \textit{Multi-Choice QA}, we map the Video LLM response to an option if the response matches with the complete option (e.g., "A. melting") or the option indicator (e.g., "A"). For \textit{Caption Matching}, we match the Video LLM response with the complete option (e.g., "Caption A: Ice cream is melting."), the option sentence (e.g., "Ice cream is melting.") or the option indicator (e.g., "Caption A"). In terms of \textit{Yes/No QA}, we check if the Video LLM response starts with "yes" or "no". Once the Video LLM response has been mapped to a specific option, we proceed by comparing that chosen option with the ground-truth answer to assess the correctness of the response.

\subsection{LLM-based Evaluation}
If a Video LLM response fails to match any of the options, we resort to LLM-based evaluation. For \textit{Multi-Choice QA}, \textit{Yes/No QA} and \textit{Caption Matching}, we present task instruction, Video LLM response and ground-truth answer to ChatGPT and prompt it to determine whether the response is correct. The detailed prompts are shown in Table \ref{tab:eval_prompt_x}.

Regarding the task of \textit{Caption Generation}, we engage ChatGPT to tackle the corresponding \textit{Multi-Choice QA} task, using the caption generated by Video LLM contextual reference, as described in Section \ref{sec:auto_eval}. To enhance the accuracy of ChatGPT in answering the \textit{Multi-Choice} questions, we present it with several in-context learning examples and prompt it to generate an extra reasoning step prior to obtaining the final answer. Table \ref{tab:eval_prompt_captioning} presents a clear illustration of the prompt structure used in this process.

\begin{figure}[t]
\centering
\includegraphics[width=1.\linewidth]{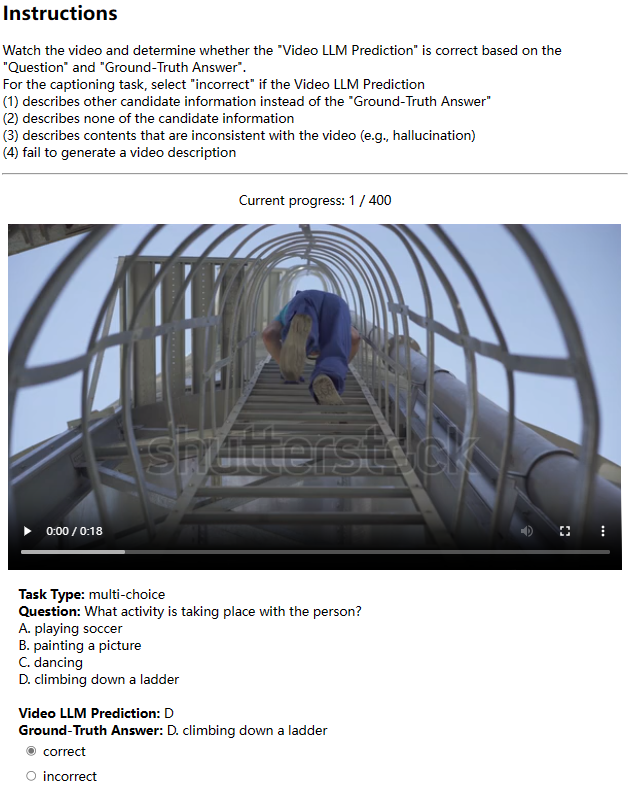}
\caption{Screenshot of the human evaluation interface.}
\label{fig:human_eval_screenshot}
\end{figure}

\subsection{Human Evaluation}
\label{app:human_eval}
To conduct human evaluation, we randomly sample 400 responses from SPHINX-v2 and Video-LLaVA, ensuring that each of the four tasks contains an equal share of 100 samples. The video, MLLM response, task instruction and ground-truth answer are presented to three human annotators, who then assign binary labels indicating the correctness of the MLLM response. For the \textit{Caption Generation} task, an MLLM response is deemed as incorrect if it (1) describes other candidate information instead of the "Ground-Truth Answer", (2) describes none of the candidate information, (3) describes contents that are inconsistent with the video (e.g., hallucination), or (4) fail to generate a video description. Figure \ref{fig:human_eval_screenshot} illustrates the interface used in our human evaluations.

\section{More Details of Evaluated Models}
\label{app:model_detail}

\subsection{Model Architecture}

We evaluate the performance of eight Video LLMs and three Image LLMs on TempCompass. All the evaluated models follow the prevalent MLLM paradigm and contain three primary components: a visual encoder, a vision-language connector, and an LLM. 
The details of these methods are as follows. 

\noindent\textbf{Video-LLaMA} \citep{Video-LLaMA} employs the same visual encoder as used in BLIP-2 \citep{BLIP-2} (ViT \citep{ViT} + Q-Former) and introduces a trainable video Q-Former to aggregate the representations of individual frames. Both the vision encoder and the LLM are frozen during training. We choose ``Video-LLaMA-2-13B'' for evaluation which is based on LLaMA-2-13B \citep{LLaMA-2}. 

\noindent\textbf{Video-ChatGPT} \citep{Video-ChatGPT} proposes to use spatial pooling and temporal pooling to aggregate frame features from a frozen image encoder (CLIP-ViT-L/14 \citep{CLIP}). A single linear layer is utilized to connect the pooled features to a frozen LLM (Vicuna-v1.1-7B \citep{Vicuna}). Unlike most MLLMs, Video-ChatGPT only performs single-stage instruction tuning on video-text data. 

\noindent\textbf{Valley} \citep{Valley} uses a similar pooling strategy as Video-ChatGPT and further incorporates a temporal modeling module into the vision encoder. In Valley, the LLM parameters are also fine-tuned during instruction tuning to achieve stronger performance. Our evaluation is carried out on ``Valley2-7b'' with LLaMA-2-7B as the base LLM.

\begin{table*}[t!]
    \centering
    \begin{tabular}{lcccc}
         \toprule
          & \multicolumn{2}{c}{Frame sampling} & \multicolumn{2}{c}{LLM decoding}  \\
          \cmidrule(lr){2-3} \cmidrule(lr){4-5}
          & strategy & \# frame & strategy & parameter \\
         \midrule
            \textcolor{lightgray}{\textit{Image LLM}} & & & & \\
			\quad LLaVA-1.5 (13B) & Middle frame & 1 & Random & $T=0.7$ \\
			\quad SPHINX-v2 (13B) & Middle frame & 1 & Top-$p$ & $T=0.9$, $p=0.8$ \\
			\quad Qwen-VL-Chat (7B) & Middle frame & 1 & Top-$p$ & $p=0.3$ \\
         \midrule
            \textcolor{lightgray}{\textit{Video LLM}} & & & & \\
			\quad Video-LLaVA (7B) & Uniform & 8 & Random & $T=0.1$ \\
			\quad LLaMA-VID (7B) & 1fps & variable & Random & $T=1.0$ \\
			\quad mPLUG-Owl (7B) & Uniform & 8 & Top-$k$ & $k=5$ \\
			\quad PandaGPT (13B) & See \citet{ImageBind} & 10 & Top-$p$ & $p=0.8$ \\
			\quad Valley (7B) & Uniform & 8 & \multicolumn{2}{c}{Greedy} \\
			\quad VideoChat2 (7B) & Uniform & 16 & \multicolumn{2}{c}{Greedy} \\
			\quad Video-ChatGPT (7B) & Uniform & 100 & Random & $T=0.2$ \\
			\quad Video-LLaMA (13B) & Uniform & 8 & Top-$p$ & $p=0.8$ \\
         \bottomrule
    \end{tabular}
    \caption{Inference settings for the evaluated MLLMs.}
    \label{tab:inference_settings}
\end{table*}

\noindent\textbf{VideoChat2} \citep{MVBench} adopts UMT-L \citep{UMT} as the vision encoder, Vicuna-v0-7B as the LLM, and utilizes a Q-Former to connect both modalities. It follows a progressive three-stage training strategy including vision-language alignment, vision-language connection, and instruction tuning. 

\noindent\textbf{mPLUG-Owl} \citep{mPLUG-Owl} proposes to use a visual abstractor similar to the Q-Former to connect the vision encoder and the LLM. It incorporates both language-only data and multimodal data into the instruction tuning procedure. Its video version, ``mPLUG-Owl-video-7B'', uses LLaMA-7B \citep{LLaMA} as the LLM and introduces additional temporal query tokens into the visual abstractor for temporal modeling.

\noindent\textbf{PandaGPT} \citep{PandaGPT} adopts ImageBind \citep{ImageBind} as the visual encoder which is pre-trained for multi-modal alignment. Similar to LLaVA \citep{LLaVA}, the vision-language connector consists only of a linear projection. Only the projection and additional LoRA \citep{LoRA} weights on LLM attention modules are updated during single-stage instruction tuning. We test ``pandagpt-13b-max-len-400'' on our dataset, which uses Vicuna-v0-13B as the LLM. 

\noindent\textbf{Video-LLaVA} \citep{Video-LLaVA} uses LanguageBind \citep{LanguageBind} to encode visual inputs and a linear layer to project visual features into the LLM space. LanguageBind and the LLM (Vicuna-v1.5-7B) are both frozen during the two-stage training. 

\noindent\textbf{LLaMA-VID} \citep{LLaMA-VID} represents each frame with two tokens, a text-guided context token and a visual content token, which significantly reduces computational cost when increasing the number of sampled frames. We evaluate the performance of ``llama-vid-7b-full-224-video-fps-1'', which is based on EVA-ViT-G \citep{EVA} and Vicuna-v1.5-7B.

\noindent\textbf{LLaVA-1.5} \citep{LLaVA-v1.5} is an Image LLM built upon the pioneering framework of LLaVA. It replaces the original linear connector with an MLP and includes additional training data to enhance its capabilities. The version we test is ``LLaVA-1.5-13B'', which adopts Vicuna-v1.5-13B as the LLM. 

\noindent\textbf{SPHINX} \citep{SPHINX} achieves high performance on many Image LLM benchmarks by mixing visual embeddings from various vision backbones including ViT, ConvNeXt \citep{ConvNeXt}, DINOv2 \citep{DINOv2}, and Q-Former. We evaluate ``SPHINX-v2'' on our benchmark which is built upon LLaMA-2-13B.

\noindent\textbf{Qwen-VL-Chat} \citep{QWen-VL-Chat} utilizes a single-layer cross-attention module with learnable query embeddings as the vision-language connector. It undergoes extensive vision-language pre-training before fine-tuned on multi-modal instruction data. The LLM of Qwen-VL-Chat is initialized with Qwen-7B \citep{QWen}. 

\subsection{Inference Settings}
\label{app:infer_setting}
Table \ref{tab:inference_settings} shows the detailed inference settings for the MLLMs. 
The frame sampling strategies of Video LLMs and the LLM decoding strategies of all the evaluated MLLMs are determined according to the recommended inference script in their corresponding codebases. For Image LLMs, we extract the middle frame of each video as the visual input to these models. 

Inspired by \cite{MVBench}, we append answer prompts to the task instructions to guide MLLMs generating responses in the desired formats. For \textit{Multi-Choice QA} and \textit{Caption Matching}, we use ``\textit{Best Option:}'' Regarding VideoChat2, an additional left bracket is appended (i.e., ``\textit{Best Option: (}'') following the original paper \cite{MVBench}. In the case of \textit{Yes/No QA}, we introduce the prompt of ``\textit{Please answer yes or no: }''. Lastly, for \textit{Caption Generation}, we use ``\textit{Generated Caption:}''. Unless otherwise specified, all the results in this paper are obtained using the above answer prompts.

\section{More Experimental Results}
\label{app:more_results}
\subsection{Results on Fine-Grained Temporal Aspects}
\label{app:quantative_results}
Table \ref{tab:eval_results} summarizes the evaluation results on all fine-grained temporal aspects.

\subsection{Effect of Answer Prompt}
\label{app:answer_prompt}
Table \ref{tab:effect_ans_prompt} reports the results on \textit{Multi-Choice QA} and \textit{Caption Matching} when using two different answer prompts, i.e., ``\textit{Best Option:}" and ``\textit{Please directly give the best option:}". The following observations can be derived: (1) The selection of answer prompt has a non-negligible impact on the match rate. The latter answer prompt, which is more detailed, can substantially increase the match rate for most MLLMs that already achieve >30\% match rate using the former answer prompt, on both two tasks. By contrast, the match rate of VideoChat2 significantly drops from near 100\% to near 0\%. This reveals that while VideoChat2 can respond in desired format (i.e., directly selecting an option) by identifying the left-bracket, it is not robust to the variation of answer prompts. (2) Compared with match rate, the accuracy is relatively insensitive to the chance of answer prompt. For instance, the change in accuracy of Video-LLaVA and Qwen-VL-Chat on \textit{Multi-Choice QA} is less than 1\%, despite a 50\%~60\% increase of match rate.

\begin{table*}[t]
    \centering
    \small
    \resizebox{\linewidth}{!}{
    \begin{tabular}{lcccccccccccc}
         \toprule
          & \multicolumn{2}{c}{Action} & \multicolumn{2}{c}{Direction} & \multicolumn{2}{c}{Speed} & Event Order &  \multicolumn{4}{c}{Attribute Change} & \multirow{2}{*}{Avg}\\ 
         \cmidrule(lr){2-3} \cmidrule(lr){4-5} \cmidrule(lr){6-7} \cmidrule(lr){8-8} \cmidrule(lr){9-12}
         & fine & coarse & object & camera & absolute & relative & order  & color & size & combined & other  \\
         \midrule
            \textbf{Multi-Choice QA} &&&&&&&&& \\
         \midrule
            \textcolor{lightgray}{\textit{Baseline}} &&&&&&&&& \\
			\quad Human & \multicolumn{2}{|c|}{100}  & \multicolumn{2}{|c|}{96.7} & \multicolumn{2}{|c|}{90} & 100 & \multicolumn{4}{|c|}{100} & 97.3  \\
			\quad Random & 28.4 & 29.3 & 28.3 & 26.3 & 30.6 & 33.0 & 32.2 & 30.1 & 28.9 & 26.4 & 25.9 & 29.9 \\
         \midrule
            \textcolor{lightgray}{\textit{Image LLM}} &&&&&&&&& \\
			\quad LLaVA-1.5 (13B) & 56.2 & 83.8 & 32.5 & 29.3 & 44.4 & 30.6 & 34.4 & 42.3 & 35.6 & 38.3 & \underline{50.0} & 42.8\\
                \quad SPHINX-v2 (13B) & \textbf{85.0} & \underline{94.1} & 36.2 & \textbf{39.1} & \underline{48.4} & \textbf{39.9} & 36.4 & \underline{51.3} & 40.2 & \textbf{46.7} & \underline{50.0} & \underline{50.9}\\
                \quad Qwen-VL-Chat (7B) & \underline{82.4} & 88.6 & \underline{37.4} & \underline{34.8} & 46.0 & \textbf{39.9} & \underline{40.7} & \textbf{52.6} & \underline{40.9} & \underline{43.3} & 44.4 & 50.6\\
         \midrule
            \textcolor{lightgray}{\textit{Video LLM}} &&&&&&&&& \\
			\quad Video-LLaVA (7B) & 54.9 & 83.2 & 31.7 & 33.7 & 46.0 & 33.2 & \textbf{41.4} & 39.7 & 40.2 & 35.0 & \textbf{55.6} & 44.7\\
                \quad LLaMA-VID (7B) & 34.6 & 78.4 & 30.0 & 29.3 & 30.6 & 28.5 & 30.5 & 23.1 & 25.0 & 28.3 & 38.9 & 35.3\\
                \quad mPLUG-Owl (7B) & 49.7 & 80.5 & 28.8 & 30.4 & 36.3 & 29.5 & 34.8 & 30.8 & 37.1 & 35.0 & 44.4 & 40.0\\
                \quad PandaGPT (13B) & 40.5 & 31.4 & 29.6 & 22.8 & 20.2 & \underline{35.2} & 31.8 & 30.8 & 33.3 & 25.0 & 33.3 & 31.1\\
                \quad Valley (7B) & 33.3 & 58.4 & 34.2 & 16.3 & 31.5 & 33.2 & 18.9 & 39.7 & 26.5 & 26.7 & 22.2 & 31.8\\
                \quad VideoChat2 (7B) & 80.4 & \textbf{95.1} & \textbf{39.1} & 29.3 & \textbf{54.0} & 34.2 & \underline{40.7} & \textbf{52.6} & \textbf{43.9} & \underline{43.3} & 33.3 & \textbf{51.1}\\
                \quad Video-ChatGPT (7B) & 28.8 & 62.2 & 33.7 & 26.1 & 28.2 & 28.5 & 37.1 & 26.9 & 31.1 & 35.0 & 33.3 & 35.2\\
                \quad Video-LLaMA (13B) & 40.5 & 65.4 & 26.7 & 18.5 & 28.2 & 28.0 & 32.8 & 26.9 & 25.0 & 33.3 & 44.4 & 33.9\\
         \bottomrule
         \midrule
            \textbf{Yes/No QA} &&&&&&&&& \\
         \midrule
            \textcolor{lightgray}{\textit{Baseline}} &&&&&&&&& \\
                \quad Human & \multicolumn{2}{|c|}{96.7}  & \multicolumn{2}{|c|}{83.3} & \multicolumn{2}{|c|}{96.7} & 93.3 & \multicolumn{4}{|c|}{100} & 94  \\
                \quad Random & 50.0 & 50.0 & 50.0 & 50.0 & 50.0 & 50.0 & 50.0 & 50.0 & 50.0 & 50.0 & 50.0 & 50.0 \\
             \midrule
                \textcolor{lightgray}{\textit{Image LLM}} &&&&&&&&& \\
                \quad LLaVA-1.5 (13B) & 65.4 & 82.4 & 48.3 & 50.0 & 48.1 & 49.4 & 49.5 & \underline{55.1} & 52.7 & \textbf{62.0} & 50.0 & 56.4\\
                \quad SPHINX-v2 (13B) & \underline{72.1} & \underline{84.8} & 50.6 & \underline{52.7} & 58.7 & \underline{52.6} & \textbf{54.5} & 44.1 & 51.6 & 55.0 & \underline{58.3} & \underline{59.1}\\
                \quad Qwen-VL-Chat (7B) & \textbf{74.0} & \textbf{87.6} & \underline{51.4} & 52.0 & \textbf{61.9} & \textbf{58.6} & 50.8 & 45.6 & 51.6 & 52.0 & 37.5 & \textbf{60.0}\\
        \midrule
            \textcolor{lightgray}{\textit{Video LLM}} &&&&&&&&& \\
                \quad Video-LLaVA (7B) & 58.4 & \textbf{87.6} & \underline{51.4} & \underline{52.7} & 50.8 & 50.0 & 49.2 & 52.2 & 50.0 & 53.0 & 45.8 & 56.4\\
                \quad LLaMA-VID (7B) & 53.9 & 70.6 & 48.3 & 50.0 & 53.4 & 46.8 & 48.4 & 54.4 & 50.0 & 55.0 & 54.2 & 53.0\\
                \quad mPLUG-Owl (7B) & 54.6 & 72.4 & 50.9 & 50.0 & 52.4 & 50.6 & 51.3 & \textbf{57.4} & 52.7 & 49.0 & 29.2 & 54.4\\
                \quad PandaGPT (13B) & 53.9 & 52.3 & 50.3 & 48.0 & 47.6 & \underline{52.6} & \underline{53.7} & 53.7 & 52.7 & 47.0 & \textbf{62.5} & 51.8\\
                \quad Valley (7B) & 50.2 & 64.7 & \textbf{52.3} & 51.4 & 57.7 & 49.7 & 50.3 & \textbf{57.4} & 51.1 & 52.0 & 45.8 & 53.5\\
                \quad VideoChat2 (7B) & 62.5 & 81.4 & \textbf{52.3} & \textbf{57.4} & \underline{59.3} & 50.9 & 51.3 & 50.7 & \underline{54.3} & \underline{58.0} & 50.0 & 58.0\\
                \quad Video-ChatGPT (7B) & 50.2 & 54.5 & 50.0 & 50.0 & 49.7 & 49.4 & 51.0 & 50.0 & 50.0 & 50.0 & 50.0 & 50.7\\
                \quad Video-LLaMA (13B) & 58.7 & 75.9 & 45.1 & 48.0 & 53.4 & 46.3 & 51.8 & 43.4 & \textbf{55.3} & 54.0 & 45.8 & 53.7\\
         \bottomrule
         \midrule
            \textbf{Caption Matching} &&&&&&&&& \\
         \midrule
            \textcolor{lightgray}{\textit{Baseline}} &&&&&&&&& \\
                \quad Human & \multicolumn{2}{|c|}{100}  & \multicolumn{2}{|c|}{96.7} & \multicolumn{2}{|c|}{100} & 100 & \multicolumn{4}{|c|}{100} & 99.3 \\
			\quad Random & 50.0 & 50.0 & 50.0 & 50.0 & 50.0 & 50.0 & 50.0 & 50.0 & 50.0 & 50.0 & 50.0 & 50.0\\
         \midrule
            \textcolor{lightgray}{\textit{Image LLM}} &&&&&&&&& \\
			\quad LLaVA-1.5 (13B) & \underline{82.6} & 90.8 & 48.9 & \textbf{55.6} & \underline{61.6} & 51.0 & 55.0 & 39.7 & 50.0 & \textbf{66.7} & \underline{55.6} & 59.5\\
                \quad SPHINX-v2 (13B) & \underline{82.6} & \underline{95.4} & 54.0 & 46.7 & 54.5 & 43.2 & 53.0 & 56.4 & 50.8 & \underline{63.3} & \underline{55.6} & 59.2\\
                \quad Qwen-VL-Chat (7B) & \textbf{86.1} & 94.1 & \underline{56.5} & 45.6 & 55.6 & \underline{54.7} & \textbf{60.3} & \textbf{64.1} & \underline{55.3} & 51.7 & \underline{55.6} & \underline{63.1}\\
         \midrule
            \textcolor{lightgray}{\textit{Video LLM}} &&&&&&&&& \\
			\quad Video-LLaVA (7B) & 79.2 & \textbf{96.7} & 53.2 & \textbf{55.6} & \textbf{69.7} & \textbf{57.8} & \underline{57.0} & \underline{60.3} & \textbf{56.1} & 60.0 & \textbf{61.1} & \textbf{63.7}\\
                \quad LLaMA-VID (7B) & 61.8 & 83.0 & 47.7 & 40.0 & 56.6 & 50.0 & 49.0 & 51.3 & 47.7 & 46.7 & \underline{55.6} & 53.6\\
                \quad mPLUG-Owl (7B) & 54.9 & 58.8 & 45.6 & 44.4 & 48.5 & 45.3 & 49.3 & 42.3 & 50.8 & 51.7 & \underline{55.6} & 49.3\\
                \quad PandaGPT (13B) & 54.2 & 58.8 & 50.6 & \underline{53.3} & 45.5 & 43.8 & 55.0 & 55.1 & 47.0 & 46.7 & 44.4 & 51.3\\
                \quad Valley (7B) & 16.7 & 14.4 & 23.2 & 16.7 & 27.3 & 19.3 & 28.3 & 21.8 & 22.0 & 26.7 & 22.2 & 22.0\\
                \quad VideoChat2 (7B) & 56.9 & 72.5 & \textbf{57.0} & 45.6 & 56.6 & 50.5 & 53.0 & 57.7 & 54.5 & 46.7 & \underline{55.6} & 55.6\\
                \quad Video-ChatGPT (7B) & 61.1 & 68.0 & 48.1 & 50.0 & 45.5 & 49.0 & 49.3 & 47.4 & 46.2 & 56.7 & 44.4 & 51.8\\
                \quad Video-LLaMA (13B) & 65.3 & 80.4 & 48.1 & 45.6 & 52.5 & 44.3 & 52.0 & 50.0 & 43.2 & 55.0 & \underline{55.6} & 53.5\\
         \bottomrule
         \midrule
            \textbf{Caption Generation} &&&&&&&&& \\
         \midrule
            \textcolor{lightgray}{\textit{Baseline}} &&&&&&&&& \\
                \quad Human & \multicolumn{2}{|c|}{100}  & \multicolumn{2}{|c|}{86.7} & \multicolumn{2}{|c|}{100} & 100 & \multicolumn{4}{|c|}{100} & 97.3  \\
			\quad Random & 28.3 & 29.2 & 28.8 & 27.2 & 30.8 & 33.2 & 32.1 & 29.5 & 28.8 & 26.7 & 29.2 & 30.0 \\
         \midrule
            \textcolor{lightgray}{\textit{Image LLM}} &&&&&&&&& \\
			\quad LLaVA-1.5 (13B) & \textbf{56.2} & \underline{77.9} & \textbf{36.1} & 20.8 & 25.0 & 24.6 & 33.0 & \underline{41.3} & 34.1 & 35.0 & 20.8 & \underline{38.4}\\
                \quad SPHINX-v2 (13B) & \underline{54.2} & \textbf{80.9} & 23.7 & 6.7 & 14.4 & 23.4 & 37.2 & 32.7 & 29.5 & 32.5 & 29.2 & 34.9\\
                \quad Qwen-VL-Chat (7B) & 47.4 & 77.0 & 29.4 & 23.3 & 25.0 & \textbf{32.0} & 34.8 & 28.8 & 34.7 & 30.0 & 37.5 & 37.3\\
         \midrule
            \textcolor{lightgray}{\textit{Video LLM}} &&&&&&&&& \\
			\quad Video-LLaVA (7B) & 33.3 & 67.2 & 29.7 & \underline{25.8} & 18.2 & 25.8 & \underline{38.2} & 31.7 & 36.9 & 26.2 & \underline{41.7} & 34.8\\
                \quad LLaMA-VID (7B) & 38.5 & 66.7 & 28.8 & \underline{25.8} & 18.9 & 23.4 & 35.5 & 36.5 & 35.2 & \underline{37.5} & 33.3 & 34.8\\
                \quad mPLUG-Owl (7B) & 38.0 & 54.4 & 28.8 & \textbf{26.7} & \underline{35.6} & 27.7 & 31.2 & 33.7 & \textbf{38.6} & 35.0 & 37.5 & 34.4\\
                \quad PandaGPT (13B) & 26.0 & 21.6 & 28.2 & 19.2 & 21.2 & 28.5 & 29.8 & 30.8 & 36.4 & 25.0 & 37.5 & 27.5\\
                \quad Valley (7B) & 25.0 & 24.5 & 23.7 & 11.7 & 19.7 & 23.0 & 35.8 & 31.7 & 29.0 & 25.0 & 37.5 & 26.3\\
                \quad VideoChat2 (7B) & 45.8 & 61.8 & \underline{32.9} & \underline{25.8} & \textbf{40.2} & \underline{28.9} & 34.2 & \textbf{43.3} & \underline{38.1} & \textbf{47.5} & 37.5 & \textbf{38.5}\\
                \quad Video-ChatGPT (7B) & 26.0 & 54.9 & 30.4 & 23.3 & 20.5 & 26.6 & 31.8 & 36.5 & 34.1 & 30.0 & 33.3 & 31.8\\
                \quad Video-LLaMA (13B) & 41.7 & 66.2 & 23.1 & 16.7 & 15.2 & 13.3 & \textbf{38.5} & 28.8 & 34.7 & 33.8 & \textbf{50.0} & 32.2\\
         \bottomrule
    \end{tabular}
    }
    \caption{Results of the evaluation experiments. The best and second-best MLLM results are \textbf{bold} and \underline{underlined}, respectively.}
    \label{tab:eval_results}
\end{table*}
\begin{table*}[t]
    \centering
    \small
    \resizebox{\linewidth}{!}{
        \begin{tabular}{lccccccccccc}
             \toprule
              & \multicolumn{3}{c}{Image LLM} & \multicolumn{8}{c}{Video LLM} \\ 
              \cmidrule(lr){2-4} \cmidrule(lr){5-12}
              & LLaVA-1.5 & SPHINX-v2 & Qwen-VL-Chat & V-LLaVA & LLaMA-VID & mPLUG-Owl & PandaGPT & Valley & VideoChat2 & V-ChatGPT & V-LLaMA \\
              & 13B & 13B & 7B & 7B & 7B & 7B & 13B & 7B & 7B & 7B & 13B \\
             \toprule
                \multicolumn{3}{l}{\textbf{Multi-Choice QA}}  \\
             \midrule
             \multicolumn{3}{l}{\textcolor{lightgray}{\textit{Prompt 1}}} \\
                Avg Acc & 42.8 & \underline{50.9} & 50.6 & 44.7 & 35.3 & 40.0 & 31.1 & 31.8 & \textbf{51.1} & 35.2 & 33.9 \\
                Match Rate & 84.2 & \underline{99.6} & 46.8 & 37.9 & 62.9 & 3.1 & 6.4 & 3.5 & \textbf{100.0} & 1.3 & 0.6 \\
             \multicolumn{3}{l}{\textcolor{lightgray}{\textit{Prompt 2}}} \\
                Avg Acc & 47.4 & \underline{50.6} & \textbf{51.1} & 45.6 & 38.0 & 36.4 & 34.4 & 29.6 & 42.9 & 37.7 & 31.3 \\
                Match Rate & \underline{99.9} & \textbf{100.0} & 98.5 & \textbf{100.0} & 97.0 & 13.7 & 3.9 & 0.4 & 0.0 & 0.2 & 3.3 \\
             \bottomrule
             \toprule
                \multicolumn{3}{l}{\textbf{Caption Matching}} \\
             \midrule
             \multicolumn{3}{l}{\textcolor{lightgray}{\textit{Prompt 1}}} \\
                Avg Acc & 59.5 & 59.2 & \underline{63.1} & \textbf{63.7} & 53.6 & 49.3 & 51.3 & 22.0 & 55.6 & 51.8 & 53.5 \\
                Match Rate & 91.2 & 89.3 & \underline{91.6} & 76.6 & 44.5 & 15.8 & 30.7 & 11.2 & \textbf{95.3} & 7.5 & 0.1 \\
             \multicolumn{3}{l}{\textcolor{lightgray}{\textit{Prompt 2}}} \\
                Avg Acc & \textbf{64.3} & \textbf{64.3} & \underline{64.1} & 63.3 & 56.0 & 48.5 & 51.6 & 34.6 & 53.7 & 53.7 & 54.2 \\
                Match Rate & 98.2 & \textbf{99.9} & 96.0 & \underline{99.5} & 68.3 & 63.3 & 22.5 & 3.7 & 1.5 & 16.5 & 0.5 \\
             \bottomrule
        \end{tabular}
    }
    \caption{Accuracy and match rate when using different answer prompts. Prompt 1 is ``\textit{Best Option: (}'' for VideoChat2 and ``\textit{Best Option:}'' for the remaining MLLMs. Prompt 2 is ``\textit{Please directly give the best option:}".}
    \label{tab:effect_ans_prompt}
\end{table*}

\subsection{Qualitative Results}
\label{app:more_qualitative_results}
Table \ref{tab:eval_example_1}, \ref{tab:eval_example_2}, \ref{tab:eval_example_3}, \ref{tab:eval_example_4}, \ref{tab:eval_example_5}, \ref{tab:eval_example_6}, \ref{tab:eval_example_7}, \ref{tab:eval_example_8} demonstrate examples of MLLM responses alongside our automated assessment results. We can find that: (1) It is evident that the models demonstrate a deficiency in genuine temporal perception skills in terms of \textit{speed}, \textit{direction}, \textit{event order} and \textit{attribute change}. While they manage to provide accurate answers for most questions in certain videos, their performance falters when confronted with corresponding conflicting videos. (2) The proposed automatic evaluation method is reliable. Despite the arbitrary form of MLLM responses, our method can offer accurate assessment in most cases. (3) Our LLM-based evaluation method mistakenly assesses a small portion of incorrect captions as correct (Table \ref{tab:eval_example_5}, \ref{tab:eval_example_6}, \ref{tab:eval_example_7}), which echoes with the results in Table \ref{tab:auto_eval_acc}. We find that such inaccurate evaluation is mostly caused by the failure to detect hallucinated contents in the captions. Table \ref{tab:eval_example_captioning} presents two more detailed evaluation examples with intermediate reasoning steps by ChatGPT. As we can see, ChatGPT is able to select the correct option in \textit{Multi-Choice QA}, despite the existence of hallucinated content in the generated captions, thereby leading to inaccurate assessment.

\section{Licencing and Intended Use}
Our TempCompass benchmark is under CC-BY 4.0 license. The videos and textual annotation in this work should only be used for research purposes.

\begin{table*}[t]
    \centering
    \begin{tcolorbox}    
        You will receive a {\color[HTML]{FE0000} [X]} question, the ground-truth answer and the prediction from a question answering (QA) model. Your task is to determine whether QA model prediction is correct, based on the question and ground-truth answer. If the prediction is correct, respond "Correct". If the prediction is incorrect, respond "Incorrect". \\

        {\color[HTML]{FE0000} [X]} Question:\\
        {\color[HTML]{0000FE} [question]} \\
        Ground-Truth Answer:\\
        {\color[HTML]{0000FE} [ground\_truth\_answer]}\\
        Model Prediction:\\
        {\color[HTML]{0000FE} [video\_llm\_prediction]}\\
    \end{tcolorbox}
    \caption{The prompt used to evaluate \textit{Multi-Choice QA}, \textit{Yes/No QA} and \textit{Caption Matching}, where $[X] \in \{\text{Multi-Choice}, \text{Yes/No}, \text{Caption Matching}\}$.}
    \label{tab:eval_prompt_x}
\end{table*}
\begin{table*}
    \centering
    \begin{tcolorbox}    
        You will receive a video description and a multi-choice question. Your task is to choose the correct answer and briefly explain the reason why you choose the answer. If none of the choice candidates are correct or the video description lacks enough information to answer the question, just answer "None of the choices are correct". Please organize your response in this format:\\
        ```\\
        Reasoning: [Your reason to obtain the answer]\\
        Answer: [Your answer]\\
        ```\\

        Here are some examples of video description, multi-choice question and the expected answer:\\
        ```\\
        Video Description: A person is palying football.\\
        Multi-Choice Question:\\
        What is the person doing in the video?\\
        A. cooking
        B. palying football
        C. playing basketball
        D. reading book\\
        Reasoning: The video description mentions that the person is playing football.\\
        Answer: B. palying football\\
        
        Video Description: A bird is flying clockwise.\\
        Multi-Choice Question:\\
        In which direction is the bird flying?
        A. backwark
        B. counter-clockwise
        C. clockwise
        D. downward\\
        Reasoning: The video description mentions that the bird is flying clockwise\\
        Answer: C. clockwise\\
        
        Video Description: An air balloon is inflating.\\
        Multi-Choice Question:\\
        What is happening to the air balloon?
        A. exploding
        B. getting smaller
        C. flying\\
        Reasoning: The video description mentions that the air balloon is inflating, while none of the choices can be explained as inflating.\\
        Answer: None of the choices are correct\\
        ```\\

        Video Description:\\
        {\color[HTML]{0000FE} [video\_llm\_prediction]}\\
        Multi-Choice Question:\\
        {\color[HTML]{0000FE} [multi\_choice\_question]}\\
        Answer:\\
    \end{tcolorbox}
    \caption{The prompt used to answer the {\color[HTML]{0000FE} [multi\_choice\_question]} using the generated video caption as context. The answer from ChatGPT is compared with the ground-truth to assess the correctness of generated caption.}
    \label{tab:eval_prompt_captioning}
\end{table*}
\begin{table*}[t]
    \centering
    \begin{tcolorbox}    
        You will receive a piece of meta-information in the form of JSON dictionary. The meta-information consists of a "subject" and a temporal dimension (related to "action", "speed", "direction", "order" or "attribute change"). Your task is to generate 5 multi-choice questions and a correct answer based on the meta-information. Ensure that the 5 questions are diverse in language, diverse in format and diverse in the set of choices.\\
        Ensure that the question can be answered from the given meta-information. \\

        Here are some examples of meta-information and generated questions: \\
        ```\\
        {\color[HTML]{EF0000} [in\_context\_examples]} \\
        ```\\

        Meta-information: {\color[HTML]{0000FE} [meta\_information]} \\
        Generate 5 multi-choice questions and correct answers related to "{\color[HTML]{0000FE} [temporal\_aspect]}". Generate the correct answer after every generated question. Separate the questions with the string "[SEP]" and don't list the number of questions.
    \end{tcolorbox}
    \begin{tcolorbox}
        {\color[HTML]{EF0000} [in\_context\_examples]}:\\
        Meta-information: \{"subject": "boy", "action": "playing basketball"\}\\
        Multi-Choice Question:\\
        What is the boy doing in the video?\\
        A. cooking
        B. singing
        C. playing basketball\\
        Correct Answer: C. playing basketball\\
        
        Meta-information: \{"subject": "entire video", "speed": "normal speed"\}\\
        Multi-Choice Question:\\
        What is the speed of the video?
        A. normal speed
        B. time-lapse
        C. slow motion\\
        Correct Answer: A. normal speed\\
        
        Meta-information: \{"subject": "car", "direction": "turning left"\}\\
        Multi-Choice Question:\\
        In which direction is the car driving?
        A. straightforward
        B. leftwards
        C. rightwards\\
        Correct Answer: B. leftwards\\
        
        Meta-information: \{"subject": "girl", "event1": "dressing up", "event2": "leaving the room"\}\\
        Multi-Choice Question:\\
        What is the girl doing?
        A. dressing up and then leaving the room
        B. entering the room and dressing up
        C. turning off clothes and then leaving the room
        D. entering the room and then turning off clothes\\
        Correct Answer: A. dressing up and then leaving the room\\
        
        Meta-information: \{"subject": "balloon", "attribute\_change": "exploding"\}\\
        Multi-choice Question:\\
        What is happening to the balloon?
        A. shrinking
        B. stay in the same shape
        C. exploding\\
        Correct Answer: C. exploding\\
    \end{tcolorbox}
    \caption{The prompt used to generate \textit{Multi-Choice QA} instructions. {\color[HTML]{0000FE} [meta\_information]} and {\color[HTML]{0000FE} [temporal\_aspect]} are dependent on the given meta-information. {\color[HTML]{EF0000} [in\_context\_examples]} are fixed for all \textit{Multi-Choice QA} instructions.}
    \label{tab:instrut_gen_multi_choice}
\end{table*}
\begin{table*}[t]
    \centering
    \begin{tcolorbox}    
        You will receive information about several multi-choice questions in the form of JSON dictionary. The dictionaries consist of a "question" that describes the question and choices and an "answer" that describes the correct answer. Your task is to generate a positive question and a negative question for each multi-choice question. The positive questions, which are related to the correct "answer" of multi-choice question, should be answered with "yes". The negative questions, which are related to other choices except for the correct "answer", should be answered with "no". \\Ensure that the generated questions are diverse in language and do NOT fabricate information that does not exist in the given multi-choice question.\\

        Here is an example of multi-choice questions and generated positive and negative questions: \\
        ```\\
        Multi-Choice Questions:\\
        \{"question": "What is the person doing? A. singing B. cooking C. sleeping", "answer": "B. cooking"\}\\
        \{"question": "What is the primary action of the person? A. playing football B. cooking C. sleeping", "answer": "B. cooking"\}\\
        \{"question": "Which of the following actions best describes the person? A. singing B. cooking C. drinking tea", "answer": "B. cooking"\}\\
        Positive Questions:\\
        \{"question": "Is the person cooking?", "answer": "yes"\}\\
        \{"question": "Is the primary action of the person about cooking?", "answer": "yes"\}\\
        \{"question": "Is cooking best describes the person's action?", "answer": "yes"\}\\
        Negative Questions:\\
        \{"question": "Is the person sleeping?", "answer": "no"\}\\
        \{"question": "Is the primary action of the person about playing football?", "answer": "no"\}\\
        \{"question": "Is drinking tea best describes the person's action?", "answer": "no"\}\\
        ```\\

        Multi-Choice Questions:\\
        {\color[HTML]{0000FE} [multi\_choice\_questions]} \\
        Generate the positive and negative questions in JSON format as shown in the above example:
    \end{tcolorbox}
    \caption{The prompt used to generate \textit{Yes/No QA} instructions. {\color[HTML]{0000FE} [multi\_choice\_questions]} are generated by ChatGPT and rectified by humans.}
    \label{tab:instrut_gen_yes_no}
\end{table*}
\begin{table*}[t]
    \centering
    \begin{tcolorbox}    
        \textbf{Prompt to Generate True/False Captions}\\
        You will receive information about two multi-choice questions in the form of JSON dictionary. The dictionary consists of a "question" that describes the question and choices and an "answer" that describes the correct answer. Your task is to generate 1 true caption and 3 false captions. The true caption describes the correct "answer" of multi-choice question. The false captions describe other choices except for the correct "answer". \\
        Ensure that the generated captions are diverse in language and do NOT fabricate information that does not exist in the given multi-choice questions.\\

        Here is an example of multi-choice questions and generated true and false captions:\\
        ```\\
        Multi-Choice Questions:
        \{"question": "What is the person doing? A. singing B. cooking C. sleeping", "answer": "B. cooking"\}\\
        \{"question": "What is the action shown in the video? A. drawing B. cooking C. reading", "answer": "B. cooking"\}\\
        True Caption:\\
        A person is cooking.\\
        False Captions:\\
        A person is sleeping.\\
        A video showing a person singing.\\
        The person is reading.\\
        ```\\

        Multi-Choice Questions:\\
        {\color[HTML]{0000FE} [multi\_choice\_questions]} \\
        Generate the true and false captions, do NOT show the multi-choice question in your response: \\

        \textbf{Instruction Templates for Caption Matching}\\
        - Which caption matches the video better? Caption A: {\color[HTML]{0000FE} [caption\_a]} Caption B: {\color[HTML]{0000FE} [caption\_b]} \\
        - Which description is a more suitable match for the video? Option 1: {\color[HTML]{0000FE} [caption\_a]} Option 2: {\color[HTML]{0000FE} [caption\_b]}\\
        - Which sentence better captures the essence of the video? Sentence A: {\color[HTML]{0000FE} [caption\_a]} Sentence B: {\color[HTML]{0000FE} [caption\_b]}\\
    \end{tcolorbox}
     \caption{The prompt used to generate Ture/False captionis (upper) and the instruction templates for \textit{Caption Matching} (lower). True and False captions are randomly inserted into {\color[HTML]{0000FE} [caption\_a]} or {\color[HTML]{0000FE} [caption\_b]}. {\color[HTML]{0000FE} [multi\_choice\_questions]} are generated by ChatGPT and rectified by humans.}
    \label{tab:instrut_gen_caption_matching}
\end{table*}
\begin{table*}[t]
    \centering
    \begin{tcolorbox}  
        \textbf{Instruction Template for Caption Generation}: \\
        {\color[HTML]{EF0000} [task\_description]}\\
        Information A: \{”subject”: {\color[HTML]{0000EF} [subject]}, “{\color[HTML]{0000EF} [temporal\_aspect]}”: {\color[HTML]{0000EF} [option\_a]}\} \\
        Information B: \{”subject”: {\color[HTML]{0000EF} [subject]}, “{\color[HTML]{0000EF} [temporal\_aspect]}”: {\color[HTML]{0000EF} [option\_b]}\} \\
        Information C: \{”subject”: {\color[HTML]{0000EF} [subject]}, “{\color[HTML]{0000EF} [temporal\_aspect]}”: {\color[HTML]{0000EF} [option\_c]}\} \\
        Information D: \{”subject”: {\color[HTML]{0000EF} [subject]}, “{\color[HTML]{0000EF} [temporal\_aspect]}”: {\color[HTML]{0000EF} [option\_d]}\} \\

        \textbf{Templates for }{\color[HTML]{EF0000} [task\_description]}:\\
        - You will be presented with a video and several pieces of information. One piece of information is consistent with the video while the others are not. Please identify the information that consistent with the video and generate a video caption accordingly.\\
        - A video and multiple pieces of information will be provided to you. One of these pieces of information matches the content of the video, while the remaining ones do not. Your objective is to pinpoint the information that is in harmony with the video and craft a suitable video caption.\\
        - You will be presented with a video and several pieces of information. One piece of information is consistent with the video while the others are not. Please identify the information that consistent with the video and generate a video caption accordingly. Ensure that the generated video caption is brief.\\
        - A video and multiple pieces of information will be provided to you. One of these pieces of information matches the content of the video, while the remaining ones do not. Your objective is to pinpoint the information that is in harmony with the video and craft a suitable video caption. Ensure that the generated video caption is brief.\\
    \end{tcolorbox}
    \caption{\textit{Caption Generation} instruction templates. {\color[HTML]{0000EF} [subject]} and {\color[HTML]{0000EF} [temporal\_aspect]} are obtained from the meta-information. The {\color[HTML]{0000EF} [options]} are derived from the \textit{Multi-Choice QA} instructions. Every {\color[HTML]{EF0000} [task\_description]} template will be combined with the candidate information to construct different task instructions.}
    \label{tab:instrut_gen_captioning}
\end{table*}
\begin{table*}
    \caption{One data example in TempCompass. For each task type, we collect multiple instructions. Due to space limitation, only one instruction is shown for the caption generation task.}
    \label{tab:data_example_1}
    \centering
    \begin{tcolorbox}    
        \includegraphics[width=0.7\textwidth]{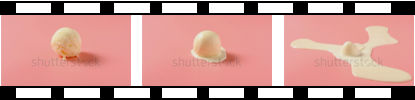} \\
        \textbf{Meta-Information}: \\
        \{"subject":"ice cream", "attribute change":"melting", "fg-aspect":"size \& shape change"\} \\
        
        \textbf{Static Contents}: food \\
        
        \textbf{Multi-Choice QA}: \\
        - What is happening to the ice cream? A. melting B. freezing C. evaporating D. solidifying \\
        - How is the ice cream changing? A. changing color B. being eaten out C. turning into liquid D. solidifying into a ball \\
        - Which term best describes the state change of the ice cream? A. evaporation B. solidifying C. melting \\

        \textbf{Yes/No QA}: \\
        - Is the ice cream melting? \\
        - Is the ice cream turning into liquid? \\
        - Is the ice cream freezing? \\
        - Is the ice cream evaporating? \\
        
        \textbf{Caption Matching}: \\
        - Which caption matches the video better? Caption A: The ice cream is melting. Caption B: The ice cream is freezing.\\
        - Which description is a more suitable match for the video? Option 1: The ice cream is evaporating. Option 2: The ice cream is melting.\\
        - Which sentence better captures the essence of the video? Sentence A: The ice cream is melting. Sentence B: The ice cream is solidifying.\\
        
        \textbf{Caption Generation}: \\
        - You will be presented with a video and several pieces of information. One piece of information is consistent with the video while the others are not. Please identify the information that consistent with the video and generate a video caption accordingly. \\
        \hspace*{1em} Information A: \{'subject':'ice cream', 'attribute\_change':'melting'\}\\
        \hspace*{1em} Information B: \{'subject':'ice cream', 'attribute\_change':'freezing'\}\\
        \hspace*{1em} Information C: \{'subject':'ice cream', 'attribute\_change':'evaporating'\}\\
        \hspace*{1em} Information D: \{'subject':'ice cream', 'attribute\_change':'solidifying'\}
        
    \end{tcolorbox}
\end{table*}
\begin{table*}[t]
    \centering
    \begin{tcolorbox}    
        \includegraphics[width=0.7\textwidth]{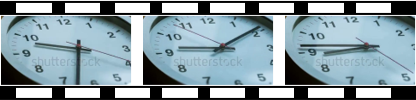} \\
        \textbf{Meta-Information}: \\
        \{"subject":"clock hands", "direction":"moving counter-clockwise", "fg-aspect":"object direction"\} \\
        
        \textbf{Static Contents}: artifacts \\
        
        \textbf{Multi-Choice QA}: \\
        - What is the direction in which the clock hands are moving? A. counterclockwise B. clockwise C. stationary \\
        - How are the clock hands changing their orientation? A. standing still B. moving clockwise C. moving anti-clockwise \\
        - In what direction are the clock hands moving? A. clockwise B. alternating between clockwise and counterclockwise C. counterclockwise \\

        \textbf{Yes/No QA}: \\
        - Are the clock hands moving counterclockwise? \\
        - Are the clock hands moving clockwise? \\
        
        \textbf{Caption Matching}: \\
        - Which description is a more suitable match for the video? Option 1: The clock hands are moving clockwise. Option 2: The clock hands are moving counterclockwise.\\
        - Which sentence better captures the essence of the video? Sentence A: The clock hands are moving counterclockwise. Sentence B: The clock hands are moving clockwise.\\
        - Which caption matches the video better? Caption A: The clock hands are rotating counterclockwise. Caption B: The clock hands are rotating clockwise.\\
        
        \textbf{Caption Generation}: \\
        - You will be presented with a video and several pieces of information. One piece of information is consistent with the video while the others are not. Please identify the information that consistent with the video and generate a video caption accordingly.\\
        \hspace*{1em} Information A: \{'subject':'clock hands', 'direction':'counterclockwise'\}\\
        \hspace*{1em} Information B: \{'subject':'clock hands', 'direction':'clockwise'\}\\
        \hspace*{1em} Information C: \{'subject':'clock hands', 'direction':'stationary'\}
        
    \end{tcolorbox}
    \caption{One data example in TempCompass. For each task type, we collect multiple instructions. Due to space limitation, only one instruction is shown for the caption generation task.}
    \label{tab:data_example_2}
\end{table*}
\begin{table*}[t]
    \centering
    \begin{tcolorbox}    
        \includegraphics[width=0.7\textwidth]{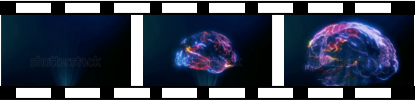} \\
        \textbf{Meta-Information}: 
        \{"subject":"entire video", "direction":"zoom into a 3d digital brain", "fg-aspect":"camera direction"\} \\
        
        \textbf{Static Contents}: abstract \\
        
        \textbf{Multi-Choice QA}: \\
        - What is happening in the video? A. Zoom out from a 3D digital brain B. Standing still before a 3D digital brain C. Zoom into a 3D digital brain \\
        - In which direction is the camera moving in the video? A. zooming in B. downwards C. upwards D. zooming out \\
        - How would you describe the trajectory of the video's direction? A. panning right B. panning left C. zooming out D. zooming in \\

        \textbf{Yes/No QA}: \\
        - Is the camera zooming into a 3D digital brain? \\
        - Is the camera moving in the video by zooming in? \\
        - Is the camera standing still before a 3D digital brain? \\
        - Is the camera zooming out in the video? \\
        
        \textbf{Caption Matching}: \\
        - Which caption matches the video better? Caption A: The camera is standing still before a 3D digital brain. Caption B: The camera is zooming into a 3D digital brain.\\
        - Which description is a more suitable match for the video? Option 1: The camera is zooming out from a 3D digital brain. Option 2: The camera is zooming into a 3D digital brain.\\
        - Which sentence better captures the essence of the video? Sentence A: The camera is zooming into a 3D digital brain. Sentence B: The camera is moving downwards.\\
        
        \textbf{Caption Generation}: \\
        - You will be presented with a video and several pieces of information. One piece of information is consistent with the video while the others are not. Please identify the information that consistent with the video and generate a video caption accordingly.\\
        \hspace*{1em} Information A: \{'subject':'entire video', 'direction':'Zoom out from a 3D digital brain'\}\\
        \hspace*{1em} Information B: \{'subject':'entire video', 'direction':'Standing still before a 3D digital brain'\}\\
        \hspace*{1em} Information C: \{'subject':'entire video', 'direction':'Zoom into a 3D digital brain'\}
    \end{tcolorbox}
    \caption{One data example in TempCompass. For each task type, we collect multiple instructions. Due to space limitation, only one instruction is shown for the caption generation task.}
    \label{tab:data_example_3}
\end{table*}
\begin{table*}[t]
    \centering
    \begin{tcolorbox}    
        \includegraphics[width=0.7\textwidth]{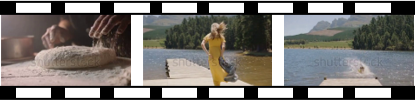} \\
        \textbf{Meta-Information}: 
        \{"subject":"entire video", "event1":"a person is kneading dough", "event2":"a girl is jumping into water"\} \\
        
        \textbf{Static Contents}: food, people, natural objects \\
        
        \textbf{Multi-Choice QA}: \\
        - What is the sequence of events in the video? A. a person is kneading dough followed by a girl jumping into water B. a girl jumping into water followed by a person kneading dough C. a girl jumping into water while a person kneading dough \\
        - What is happening first in the video? A. Both events occur at the same time B. a person is kneading dough C. a girl is jumping into water \\
        - What event occurs second in the video? A. a person is kneading dough B. a girl is jumping into water C. Both events occur at the same time \\

        \textbf{Yes/No QA}: \\
        - Is the sequence of events in the video a person kneading dough followed by a girl jumping into water? \\
        - Is the sequence of events in the video a girl jumping into water followed by a person kneading dough? \\
        
        \textbf{Caption Matching}: \\
        - Which description is a more suitable match for the video? Option 1: In the video, a person is kneading dough followed by a girl jumping into water. Option 2: A girl jumping into water while a person kneading dough.\\
        - Which sentence better captures the essence of the video? Sentence A: A girl is jumping into water, followed by a person kneading dough. Sentence B: In the video, a person is kneading dough followed by a girl jumping into water.\\
        - Which caption matches the video better? Caption A: In the video, a person is kneading dough followed by a girl jumping into water. Caption B: A person is kneading dough after a girl jumping into water.\\
        
        \textbf{Caption Generation}: \\
        - You will be presented with a video and several pieces of information. One piece of information is consistent with the video while the others are not. Please identify the information that consistent with the video and generate a video caption accordingly.\\
        \hspace*{1em} Information A: \{'subject':'entire video', 'order':'a person is kneading dough followed by a girl jumping into water'\}\\
        \hspace*{1em} Information B: \{'subject':'entire video', 'order':'a girl jumping into water followed by a person kneading dough'\}\\
        \hspace*{1em} Information C: \{'subject':'entire video', 'order':'a girl jumping into water while a person kneading dough'\}
    \end{tcolorbox}
    \caption{One data example in TempCompass. For each task type, we collect multiple instructions. Due to space limitation, only one instruction is shown for the caption generation task.}
    \label{tab:data_example_4}
\end{table*}

\begin{table*}[t]
    \centering
    \begin{tcolorbox}    
        \includegraphics[width=0.7\textwidth]{Figures/vid_pineapple.png} \\
        
        \textbf{Multi-Choice QA}: \\
        {\color[HTML]{EF0000} Q:} What is happening in the video? A. A person drops down the pineapple B. A person pushes forward the pineapple C. A person rotates the pineapple D. A person picks up the pineapple \\
        {\color[HTML]{00CC00} Ground-Truth:} D. A person picks up the pineapple\\
        \hspace*{1em} {\color[HTML]{0000EF} SPHINX-v2:} D (\checkmark)\\
        \hspace*{1em} {\color[HTML]{0000EF} Video-LLaVA:} D (\checkmark)\\
        \hspace*{1em} {\color[HTML]{0000EF} Video-ChatGPT:} The best option is to pick up the pineapple. (\checkmark)\\

        \textbf{Yes/No QA}: \\
        {\color[HTML]{EF0000} Q:} Is a person picking up the pineapple? \\
        {\color[HTML]{00CC00} Ground-Truth:} yes\\
        \hspace*{1em} {\color[HTML]{0000EF} SPHINX-v2:} Yes (\checkmark)\\
        \hspace*{1em} {\color[HTML]{0000EF} Video-LLaVA:} Yes (\checkmark)\\
        \hspace*{1em} {\color[HTML]{0000EF} Video-ChatGPT:} Yes, a person is picking up the pineapple. (\checkmark)\\
        
        \textbf{Caption Matching}: \\
        {\color[HTML]{EF0000} Q:} Which sentence better captures the essence of the video? Sentence A: A person is dropping down the pineapple. Sentence B: A person is picking up the pineapple.\\
        {\color[HTML]{00CC00} Ground-Truth:} Sentence B: A person is picking up the pineapple.\\
        \hspace*{1em} {\color[HTML]{0000EF} SPHINX-v2:} Sentence B (\checkmark)\\
        \hspace*{1em} {\color[HTML]{0000EF} Video-LLaVA:} Sentence B: A person is picking up the pineapple. (\checkmark)\\
        \hspace*{1em} {\color[HTML]{0000EF} Video-ChatGPT:} Sentence B captures the essence of the video better as it shows the person picking up the pineapple, which is the main focus of the video. (\checkmark)\\
        
        \textbf{Caption Generation}: \\
        {\color[HTML]{00CC00} Ground-Truth:} Information D: \{'subject': 'person', 'direction': 'A person picks up the pineapple'\}\\
        \hspace*{1em} {\color[HTML]{0000EF} SPHINX-v2:} A person is holding a pineapple on a pink table. (\checkmark)\\
        \hspace*{1em} {\color[HTML]{0000EF} Video-LLaVA:} A person picks up the pineapple and drops it down. (\checkmark)\\
        \hspace*{1em} {\color[HTML]{0000EF} Video-ChatGPT:} In the video, a person is seen dropping a pineapple, pushing forward a pineapple, rotating a pineapple, and picking up a pineapple. The person is seen using their hands to interact with the pineapple. The video shows the person's actions and the pineapple's movement, but the exact reason for the actions is not clear from the video. (\crossmark)\\
    \end{tcolorbox}
    \caption{An example of MLLM responses and evaluation results of the \textit{Direction} aspect. The \checkmark and \crossmark in the parentheses are assessed by our automatic evaluation method. The \textit{Caption Generation} task instruction is discarded for simplicity.}
    \label{tab:eval_example_9}
\end{table*}
\begin{table*}[t]
    \centering
    \begin{tcolorbox}    
        \includegraphics[width=0.7\textwidth]{Figures/vid_pineapple_reverse.png} \\
        
        \textbf{Multi-Choice QA}: \\
        {\color[HTML]{EF0000} Q:} What is happening in the video? A. A person drops down the pineapple B. A person pushes forward the pineapple C. A person rotates the pineapple D. A person picks up the pineapple \\
        {\color[HTML]{00CC00} Ground-Truth:} A. A person drops down the pineapple\\
        \hspace*{1em} {\color[HTML]{0000EF} SPHINX-v2:} D (\crossmark)\\
        \hspace*{1em} {\color[HTML]{0000EF} Video-LLaVA:} c (\crossmark)\\
        \hspace*{1em} {\color[HTML]{0000EF} Video-ChatGPT:} The best option would be to pick up the pineapple and place it on the table. (\crossmark)\\

        \textbf{Yes/No QA}: \\
        {\color[HTML]{EF0000} Q:} Is a person picking up the pineapple? \\
        {\color[HTML]{00CC00} Ground-Truth:} no\\
        \hspace*{1em} {\color[HTML]{0000EF} SPHINX-v2:} Yes (\crossmark)\\
        \hspace*{1em} {\color[HTML]{0000EF} Video-LLaVA:} Yes (\crossmark)\\
        \hspace*{1em} {\color[HTML]{0000EF} Video-ChatGPT:} Yes, a person is picking up the pineapple. (\crossmark)\\
        
        \textbf{Caption Matching}: \\
        {\color[HTML]{EF0000} Q:} Which sentence better captures the essence of the video? Sentence A: A person is dropping down the pineapple. Sentence B: A person is picking up the pineapple.\\
        {\color[HTML]{00CC00} Ground-Truth:} Sentence A: A person is dropping down the pineapple.\\
        \hspace*{1em} {\color[HTML]{0000EF} SPHINX-v2:} Sentence B (\crossmark)\\
        \hspace*{1em} {\color[HTML]{0000EF} Video-LLaVA:} Sentence B: A person is picking up the pineapple. (\crossmark)\\
        \hspace*{1em} {\color[HTML]{0000EF} Video-ChatGPT:} Sentence B captures the essence of the video better as it shows the person picking up the pineapple, which is the main action in the video. (\crossmark)\\
        
        \textbf{Caption Generation}: \\
        {\color[HTML]{00CC00} Ground-Truth:} Information A: \{'subject': 'person', 'direction': 'A person drops down the pineapple'\}\\
        \hspace*{1em} {\color[HTML]{0000EF} SPHINX-v2:} A person is holding a pineapple on a pink surface. (\crossmark)\\
        \hspace*{1em} {\color[HTML]{0000EF} Video-LLaVA:} A person drops down the pineapple, and then a person pushes forward the pineapple. After that, a person rotates the pineapple, and finally, a person picks up the pineapple. (\crossmark)\\
        \hspace*{1em} {\color[HTML]{0000EF} Video-ChatGPT:} In the video, a person is seen dropping a pineapple, pushing forward a pineapple, rotating a pineapple, and picking up a pineapple. The person is seen using their hands to interact with the pineapple. The video shows the person's actions and the pineapple's movement, but the exact reason for the actions is not clear from the video. (\crossmark)\\
    \end{tcolorbox}
    \caption{An example of MLLM responses and evaluation results of the \textit{Direction} aspect. The \checkmark and \crossmark in the parentheses are assessed by our automatic evaluation method. The \textit{Caption Generation} task instruction is discarded for simplicity.}
    \label{tab:eval_example_10}
\end{table*}
\begin{table*}[t]
    \centering
    \begin{tcolorbox}    
        \includegraphics[width=0.7\textwidth]{Figures/vid_ice_cream.png} \\
        
        \textbf{Multi-Choice QA}: \\
        {\color[HTML]{EF0000} Q:} What is happening to the ice cream? A. melting B. freezing C. evaporating D. solidifying \\
        {\color[HTML]{00CC00} Ground-Truth:} A. melting\\
        \hspace*{1em} {\color[HTML]{0000EF} SPHINX-v2:} A (\checkmark)\\
        \hspace*{1em} {\color[HTML]{0000EF} Video-LLaVA:} A. melting (\checkmark)\\
        \hspace*{1em} {\color[HTML]{0000EF} Video-ChatGPT:} The best option is A. melting, as the ice cream is melting in the image. (\checkmark)\\

        \textbf{Yes/No QA}: \\
        {\color[HTML]{EF0000} Q:} Is the ice cream melting? \\
        {\color[HTML]{00CC00} Ground-Truth:} yes\\
        \hspace*{1em} {\color[HTML]{0000EF} SPHINX-v2:} Yes (\checkmark)\\
        \hspace*{1em} {\color[HTML]{0000EF} Video-LLaVA:} Yes. (\checkmark)\\
        \hspace*{1em} {\color[HTML]{0000EF} Video-ChatGPT:} Yes, the ice cream is melting. (\checkmark)\\
        
        \textbf{Caption Matching}: \\
        {\color[HTML]{EF0000} Q:} Which sentence better captures the essence of the video? Sentence A: The ice cream is melting. Sentence B: The ice cream is solidifying.\\
        {\color[HTML]{00CC00} Ground-Truth:} Sentence A: The ice cream is melting.\\
        \hspace*{1em} {\color[HTML]{0000EF} SPHINX-v2:} The ice cream is melting. (\checkmark)\\
        \hspace*{1em} {\color[HTML]{0000EF} Video-LLaVA:} Sentence B: The ice cream is solidifying. (\crossmark)\\
        \hspace*{1em} {\color[HTML]{0000EF} Video-ChatGPT:} Sentence A, \"The ice cream is melting,\" captures the essence of the video more accurately as it shows the ice cream melting and dripping off the table. (\checkmark)\\
        
        \textbf{Caption Generation}: \\
        {\color[HTML]{00CC00} Ground-Truth:} Information A: \{'subject': 'ice cream', 'attribute\_change': 'melting'\}\\
        \hspace*{1em} {\color[HTML]{0000EF} SPHINX-v2:} A scoop of ice cream melting on a pink table. (\checkmark)\\
        \hspace*{1em} {\color[HTML]{0000EF} Video-LLaVA:} A scoop of ice cream melts on a pink background. (\checkmark)\\
        \hspace*{1em} {\color[HTML]{0000EF} Video-ChatGPT:} The video shows a close-up of a melting ice cream cone, which is a consistent piece of information with the video. (\checkmark)\\
    \end{tcolorbox}
    \caption{An example of MLLM responses and evaluation results of the \textit{Attribute Change} aspect. The \checkmark and \crossmark in the parentheses are assessed by our automatic evaluation method. The \textit{Caption Generation} task instruction is discarded for simplicity.}
    \label{tab:eval_example_1}
\end{table*}
\begin{table*}[t]
    \centering
    \begin{tcolorbox}    
        \includegraphics[width=0.7\textwidth]{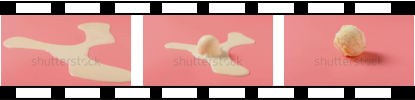} \\
        
        \textbf{Multi-Choice QA}: \\
        {\color[HTML]{EF0000} Q:} What is happening to the ice cream? A. melting B. freezing C. evaporating D. solidifying \\
        {\color[HTML]{00CC00} Ground-Truth:} D. solidifying\\
        \hspace*{1em} {\color[HTML]{0000EF} SPHINX-v2:} A (\crossmark)\\
        \hspace*{1em} {\color[HTML]{0000EF} Video-LLaVA:} A. melting (\crossmark)\\
        \hspace*{1em} {\color[HTML]{0000EF} Video-ChatGPT:} The best option is A. melting, as the ice cream is shown melting in the video.(\crossmark)\\

        \textbf{Yes/No QA}: \\
        {\color[HTML]{EF0000} Q:} Is the ice cream melting? \\
        {\color[HTML]{00CC00} Ground-Truth:} no\\
        \hspace*{1em} {\color[HTML]{0000EF} SPHINX-v2:} Yes (\crossmark)\\
        \hspace*{1em} {\color[HTML]{0000EF} Video-LLaVA:} Yes. (\crossmark)\\
        \hspace*{1em} {\color[HTML]{0000EF} Video-ChatGPT:} Yes, the ice cream is melting. (\crossmark)\\
        
        \textbf{Caption Matching}: \\
        {\color[HTML]{EF0000} Q:} Which sentence better captures the essence of the video? Sentence A: The ice cream is melting. Sentence B: The ice cream is solidifying.\\
        {\color[HTML]{00CC00} Ground-Truth:} Sentence B: The ice cream is solidifying.\\
        \hspace*{1em} {\color[HTML]{0000EF} SPHINX-v2:} The ice cream is melting. (\crossmark)\\
        \hspace*{1em} {\color[HTML]{0000EF} Video-LLaVA:} Sentence A: The ice cream is melting. (\crossmark)\\
        \hspace*{1em} {\color[HTML]{0000EF} Video-ChatGPT:} Sentence A, \"The ice cream is melting,\" captures the essence of the video better as it shows the ice cream melting and dripping off the table. (\crossmark)\\
        
        \textbf{Caption Generation}: \\
        {\color[HTML]{00CC00} Ground-Truth:} Information D: \{'subject': 'ice cream', 'attribute\_change': 'solidifying'\}\\
        \hspace*{1em} {\color[HTML]{0000EF} SPHINX-v2:} A scoop of ice cream melting on a pink table. (\crossmark)\\
        \hspace*{1em} {\color[HTML]{0000EF} Video-LLaVA:} The video shows a close-up of ice cream melting on a pink surface. The ice cream is shown melting in a circular motion, with the liquid dripping off the surface. The video emphasizes the process of melting and evaporating, with the ice cream gradually solidifying. (\crossmark)\\
        \hspace*{1em} {\color[HTML]{0000EF} Video-ChatGPT:} The video shows a close-up of an ice cream cone melting. This information is consistent with the video and is reflected in the generated caption. (\crossmark)\\
    \end{tcolorbox}
    \caption{An example of MLLM responses and evaluation result of the \textit{Attribute Change} aspect. The \checkmark and \crossmark in the parentheses are assessed by our automatic evaluation method. The \textit{Caption Generation} task instruction is discarded for simplicity.}
    \label{tab:eval_example_2}
\end{table*}
\begin{table*}[t]
    \centering
    \begin{tcolorbox}    
    \includegraphics[width=0.7\textwidth]{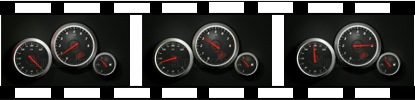} \\
    
    \textbf{Multi-Choice QA}: \\
    {\color[HTML]{EF0000} Q:} In relation to the speedometer on the left, how does the speedometer in the center appear?
    A. stationary
    B. moving slower
    C. moving faster \\
    {\color[HTML]{00CC00} Ground-Truth:} C. moving faster\\
    \hspace*{1em} {\color[HTML]{0000EF} SPHINX-v2:} C (\checkmark)\\
    \hspace*{1em} {\color[HTML]{0000EF} Video-LLaVA:} The speedometer in the center appears to be moving faster than the one on the left. (\checkmark)\\
    \hspace*{1em} {\color[HTML]{0000EF} Video-ChatGPT:} The speedometer in the center appears to be moving slower than the one on the left. (\crossmark)\\
    
    \textbf{Yes/No QA}: \\
    {\color[HTML]{EF0000} Q:} Is the speedometer in the center moving faster than the one on the left? \\
    {\color[HTML]{00CC00} Ground-Truth:} yes\\
    \hspace*{1em} {\color[HTML]{0000EF} SPHINX-v2:} yes (\checkmark)\\
    \hspace*{1em} {\color[HTML]{0000EF} Video-LLaVA:} Yes. (\checkmark)\\
    \hspace*{1em} {\color[HTML]{0000EF} Video-ChatGPT:} Yes, the speedometer in the center is moving faster than the one on the left. (\checkmark)\\

    \textbf{Caption Matching}: \\
    {\color[HTML]{EF0000} Q:} Which description is a more suitable match for the video?
    Option 1: The speedometer in the center is moving faster than the one on the left.
    Option 2: The speedometer in the center is moving slower than the one on the left. \\
    {\color[HTML]{00CC00} Ground-Truth:} Option 1: The speedometer in the center is moving faster than the one on the left.\\
    \hspace*{1em} {\color[HTML]{0000EF} SPHINX-v2:} The speedometer in the center is moving faster than the one on the right. (\crossmark)\\
    \hspace*{1em} {\color[HTML]{0000EF} Video-LLaVA:} Option 1: The speedometer in the center is moving faster than the one on the left. (\checkmark)\\
    \hspace*{1em} {\color[HTML]{0000EF} Video-ChatGPT:} Option 1 is the most suitable description for the video, as the speedometer in the center is moving faster than the one on the left. (\checkmark)\\
    
    \textbf{Caption Generation}: \\
    {\color[HTML]{00CC00} Ground-Truth:} Information C: \{'subject': 'the speedometer in the center', 'speed': 'moving faster'\}\\
    \hspace*{1em} {\color[HTML]{0000EF} SPHINX-v2:} A speedometer with the needle pointing to the number 0. (\crossmark)\\
    \hspace*{1em} {\color[HTML]{0000EF} Video-LLaVA:} The video shows a speedometer in the center of the screen, which is stationary. The speedometer is then shown moving slower, and finally moving faster. The video does not provide any additional information about the speedometer. (\crossmark)\\
    \hspace*{1em} {\color[HTML]{0000EF} Video-ChatGPT:} The video shows a speedometer in the center, with two other speedometers on either side. The speedometer in the center is stationary, while the speedometers on either side are moving slower and faster, respectively. (\crossmark)\\
    \end{tcolorbox}
    \caption{An example of MLLM responses and evaluation results of the \textit{Speed} aspect. The \checkmark and \crossmark in the parentheses are assessed by our automatic evaluation method. The \textit{Caption Generation} task instruction is discarded for simplicity.}
    \label{tab:eval_example_3}
\end{table*}
\begin{table*}[t]
    \centering
    \begin{tcolorbox}    
    \includegraphics[width=0.7\textwidth]{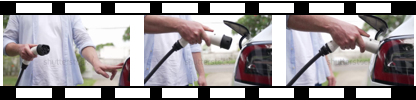} \\
    
    \textbf{Multi-Choice QA}: \\
    {\color[HTML]{EF0000} Q:} What is the speed of the video?
    A. time-lapse
    B. slow motion
    C. normal speed \\
    {\color[HTML]{00CC00} Ground-Truth:} B. slow motion\\
    \hspace*{1em} {\color[HTML]{0000EF} SPHINX-v2:} C (\crossmark)\\
    \hspace*{1em} {\color[HTML]{0000EF} Video-LLaVA:} A. time-lapse (\crossmark)\\
    \hspace*{1em} {\color[HTML]{0000EF} Video-ChatGPT:} The correct answer is A. time-lapse. (\crossmark)\\
    
    \textbf{Yes/No QA}: \\
    {\color[HTML]{EF0000} Q:} Is the video playing in slow motion? \\
    {\color[HTML]{00CC00} Ground-Truth:} yes\\
    \hspace*{1em} {\color[HTML]{0000EF} SPHINX-v2:} No (\crossmark)\\
    \hspace*{1em} {\color[HTML]{0000EF} Video-LLaVA:} Yes. (\checkmark)\\
    \hspace*{1em} {\color[HTML]{0000EF} Video-ChatGPT:} Yes, the video is playing in slow motion. (\checkmark)\\

    \textbf{Caption Matching}: \\
    {\color[HTML]{EF0000} Q:} Which description is a more suitable match for the video?
    Option 1: The video is playing in fast motion.
    Option 2: The video is playing in slow motion. \\
    {\color[HTML]{00CC00} Ground-Truth:} Option 2: The video is playing in slow motion.\\
    \hspace*{1em} {\color[HTML]{0000EF} SPHINX-v2:} The video is playing in slow motion. (\checkmark)\\
    \hspace*{1em} {\color[HTML]{0000EF} Video-LLaVA:} Option 2: The video is playing in slow motion. (\checkmark)\\
    \hspace*{1em} {\color[HTML]{0000EF} Video-ChatGPT:} The video is playing in fast motion. (\crossmark)\\
    
    \textbf{Caption Generation}: \\
    {\color[HTML]{00CC00} Ground-Truth:} Information B: \{'subject': 'entire video', 'speed': 'slow motion'\}\\
    \hspace*{1em} {\color[HTML]{0000EF} SPHINX-v2:} A man is charging his car with an electric charger. (\crossmark)\\
    \hspace*{1em} {\color[HTML]{0000EF} Video-LLaVA:} A man is seen charging his electric car in a parking lot. He uses a portable charger to charge his car. The video shows the process of charging the car and the man's actions. (\crossmark)\\
    \hspace*{1em} {\color[HTML]{0000EF} Video-ChatGPT:} Based on the provided information, the video is a time-lapse of a man pumping gas into a car. (\crossmark)\\
    \end{tcolorbox}
    \caption{An example of MLLM responses and evaluation results of the \textit{Speed} aspect. The \checkmark and \crossmark in the parentheses are assessed by our automatic evaluation method. The \textit{Caption Generation} task instruction is discarded for simplicity.}
    \label{tab:eval_example_4}
\end{table*}
\begin{table*}[t]
    \centering
    \begin{tcolorbox}    
    \includegraphics[width=0.7\textwidth]{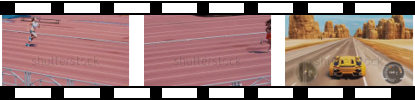} \\
    
    \textbf{Multi-Choice QA}: \\
    {\color[HTML]{EF0000} Q:} Which event happens first in the video?
    A. They happen simultaneously
    B. Cars are racing
    C. Women are running \\
    {\color[HTML]{00CC00} Ground-Truth:} C. Women are running\\
    \hspace*{1em} {\color[HTML]{0000EF} SPHINX-v2:} C (\checkmark)\\
    \hspace*{1em} {\color[HTML]{0000EF} Video-LLaVA:} Cars are racing first in the video. (\crossmark)\\
    \hspace*{1em} {\color[HTML]{0000EF} Video-ChatGPT:} The events in the video occur in the following order: cars racing, then women running. (\crossmark)\\
    
    \textbf{Yes/No QA}: \\
    {\color[HTML]{EF0000} Q:} Do the events occur in the order of women running, then cars racing? \\
    {\color[HTML]{00CC00} Ground-Truth:} yes\\
    \hspace*{1em} {\color[HTML]{0000EF} SPHINX-v2:} no (\crossmark)\\
    \hspace*{1em} {\color[HTML]{0000EF} Video-LLaVA:} Yes. (\checkmark)\\
    \hspace*{1em} {\color[HTML]{0000EF} Video-ChatGPT:} Yes, the events occur in the order of women running, then cars racing. (\checkmark)\\

    \textbf{Caption Matching}: \\
    {\color[HTML]{EF0000} Q:} Which description is a more suitable match for the video?
    Option 1: The video shows women running first, followed by cars racing.
    Option 2: Woman running and car racing happen simultaneously. \\
    {\color[HTML]{00CC00} Ground-Truth:} Option 1: The video shows women running first, followed by cars racing.\\
    \hspace*{1em} {\color[HTML]{0000EF} SPHINX-v2:} Option 2 (\crossmark)\\
    \hspace*{1em} {\color[HTML]{0000EF} Video-LLaVA:} Option 2: The video shows a woman running and cars racing simultaneously. (\crossmark)\\
    \hspace*{1em} {\color[HTML]{0000EF} Video-ChatGPT:} Option 2 is the most suitable match for the video, as it shows both a woman running and cars racing simultaneously. (\crossmark)\\
    
    \textbf{Caption Generation}: \\
    {\color[HTML]{00CC00} Ground-Truth:} Information B: \{'subject': 'entire video', 'order': 'Women running, then cars racing'\}\\
    \hspace*{1em} {\color[HTML]{0000EF} SPHINX-v2:} A woman running on a track next to a car race. (\crossmark)\\
    \hspace*{1em} {\color[HTML]{0000EF} Video-LLaVA:} The video starts with a woman running on a track, followed by several other women running. The scene then shifts to cars racing on a highway. The video then shows the women running again, followed by the cars racing. The video ends with the women running again. The video showcases the beauty of nature and the excitement of racing. The video is a perfect blend of nature and technology, making it a must-watch for all car enthusiasts. (\checkmark)\\
    \hspace*{1em} {\color[HTML]{0000EF} Video-ChatGPT:} The video shows a group of women running on a track while cars are racing nearby. The cars are racing in the background while the women are running in the foreground. This suggests that the video captures a dynamic scene where both the cars and the women are engaged in their respective activities. (\crossmark)
    \end{tcolorbox}
    \caption{An example of MLLM responses and evaluation results of the \textit{Event Order} aspect. The \checkmark and \crossmark in the parentheses are assessed by our automatic evaluation method. The \textit{Caption Generation} task instruction is discarded for simplicity.}
    \label{tab:eval_example_5}
\end{table*}
\begin{table*}[t]
    \centering
    \begin{tcolorbox}    
    \includegraphics[width=0.7\textwidth]{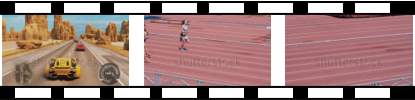} \\
    
    \textbf{Multi-Choice QA}: \\
    {\color[HTML]{EF0000} Q:} Which event happens first in the video?
    A. They happen simultaneously
    B. Cars are racing
    C. Women are running \\
    {\color[HTML]{00CC00} Ground-Truth:} B. Cars are racing\\
    \hspace*{1em} {\color[HTML]{0000EF} SPHINX-v2:} C (\crossmark)\\
    \hspace*{1em} {\color[HTML]{0000EF} Video-LLaVA:} The video shows that the women are running first, followed by the cars racing.(\crossmark)\\
    \hspace*{1em} {\color[HTML]{0000EF} Video-ChatGPT:} The events in the video happen simultaneously. The women are running on the road, and at the same time, the cars are racing on the road. (\crossmark)\\
    
    \textbf{Yes/No QA}: \\
    {\color[HTML]{EF0000} Q:} Do the events occur in the order of cars racing, then women running? \\
    {\color[HTML]{00CC00} Ground-Truth:} yes\\
    \hspace*{1em} {\color[HTML]{0000EF} SPHINX-v2:} no (\crossmark)\\
    \hspace*{1em} {\color[HTML]{0000EF} Video-LLaVA:} Yes (\checkmark)\\
    \hspace*{1em} {\color[HTML]{0000EF} Video-ChatGPT:} Yes. (\checkmark)\\

    \textbf{Caption Matching}: \\
    {\color[HTML]{EF0000} Q:} Which caption matches the video better?
    Caption A: The cars are racing first, followed by women running.
    Caption B: Woman running and car racing happen simultaneously. \\
    {\color[HTML]{00CC00} Ground-Truth:} Caption A: The cars are racing first, followed by women running.\\
    \hspace*{1em} {\color[HTML]{0000EF} SPHINX-v2:} Caption A (\checkmark)\\
    \hspace*{1em} {\color[HTML]{0000EF} Video-LLaVA:} Caption B matches the video better as it shows the women running and the cars racing simultaneously. (\crossmark)\\
    \hspace*{1em} {\color[HTML]{0000EF} Video-ChatGPT:} Caption A is the best option as it accurately describes the events in the video. The cars are racing first, followed by women running. (\checkmark)\\
    
    \textbf{Caption Generation}: \\
    {\color[HTML]{00CC00} Ground-Truth:} Information C: \{'subject': 'entire video', 'order': 'Cars racing, then women running'\}\\
    \hspace*{1em} {\color[HTML]{0000EF} SPHINX-v2:} The women are running and cars are racing. (\crossmark)\\
    \hspace*{1em} {\color[HTML]{0000EF} Video-LLaVA:} The video starts with a yellow sports car driving down a highway, followed by a group of women running on a track. The women are seen running in slow motion, and the video captures their movements in detail. The scene then shifts to a group of cars racing down the same highway, with the women running alongside them. The video captures the excitement and energy of both the car race and the women's run, showcasing the beauty of nature and the thrill of competition. (\checkmark)\\
    \hspace*{1em} {\color[HTML]{0000EF} Video-ChatGPT:} The information that is consistent with the video is that there are women running and cars racing in the video. The video shows a group of women running on a road, and at the same time, cars are racing around them. This suggests that the video captures a dynamic and exciting moment where both the women and the cars are engaged in their respective activities. (\crossmark)
    \end{tcolorbox}
    \caption{An example of MLLM responses and evaluation results of the \textit{Event Order} aspect. The \checkmark and \crossmark in the parentheses are assessed by our automatic evaluation method. The \textit{Caption Generation} task instruction is discarded for simplicity.}
    \label{tab:eval_example_6}
\end{table*}
\begin{table*}[t]
    \centering
    \begin{tcolorbox}    
    \includegraphics[width=0.7\textwidth]{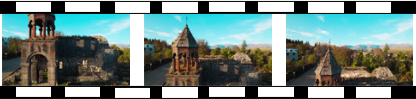} \\
    
    \textbf{Multi-Choice QA}: \\
    {\color[HTML]{EF0000} Q:} Which direction is the camera moving before the stone building?
    A. rightwards
    B. upwards
    C. leftwards
    D. downwards \\
    {\color[HTML]{00CC00} Ground-Truth:} B. upwards\\
    \hspace*{1em} {\color[HTML]{0000EF} SPHINX-v2:} B (\checkmark)\\
    \hspace*{1em} {\color[HTML]{0000EF} Video-LLaVA:} The camera is moving upwards before the stone building. (\checkmark)\\
    \hspace*{1em} {\color[HTML]{0000EF} Video-ChatGPT:} The camera is moving rightwards before the stone building. (\crossmark)\\
    
    \textbf{Yes/No QA}: \\
    {\color[HTML]{EF0000} Q:} Is the camera moving upwards before the stone building? \\
    {\color[HTML]{00CC00} Ground-Truth:} yes\\
    \hspace*{1em} {\color[HTML]{0000EF} SPHINX-v2:} Yes (\checkmark)\\
    \hspace*{1em} {\color[HTML]{0000EF} Video-LLaVA:} Yes. (\checkmark)\\
    \hspace*{1em} {\color[HTML]{0000EF} Video-ChatGPT:} Yes, the camera is moving upwards towards the stone building. (\checkmark)\\
    
    \textbf{Caption Matching}: \\
    {\color[HTML]{EF0000} Q:} Which caption matches the video better?
    Caption A: The camera is moving downwards before the stone building.
    Caption B: The movement in the video is upwards. \\
    {\color[HTML]{00CC00} Ground-Truth:} Caption B: The movement in the video is upwards.\\
    \hspace*{1em} {\color[HTML]{0000EF} SPHINX-v2:} Caption A (\crossmark)\\
    \hspace*{1em} {\color[HTML]{0000EF} Video-LLaVA:} Caption B: The movement in the video is upwards. (\checkmark)\\
    \hspace*{1em} {\color[HTML]{0000EF} Video-ChatGPT:} Caption A, which states that the camera is moving downwards before the stone building, is the best match for the video. (\crossmark)\\
    
    \textbf{Caption Generation}: \\
    {\color[HTML]{00CC00} Ground-Truth:} Information B: \{'subject': 'entire video', 'direction': 'upwards'\}\\
    \hspace*{1em} {\color[HTML]{0000EF} SPHINX-v2:} The video is a great resource for understanding the subject matter. (\crossmark)\\
    \hspace*{1em} {\color[HTML]{0000EF} Video-LLaVA:} The video showcases a beautiful and historic location with a clock tower and a church. The camera pans around the area, capturing the stunning architecture and scenery. The video also includes a shot of a mountain in the background, adding to the overall beauty of the location. (\crossmark)\\
    \hspace*{1em} {\color[HTML]{0000EF} Video-ChatGPT:} The video features a large building with a clock tower, which is located in a mountainous area. The building is surrounded by a forest, and the clock tower is situated on top of a hill. The video also shows a person walking towards the building, and a car driving by. The building appears to be old and has a rustic appearance. The clock tower is visible in the background, and the person is walking towards it. The video is shot from a distance, capturing the overall landscape and architecture of the area. (\checkmark)\\
    \end{tcolorbox}
    \caption{An example of MLLM responses and evaluation results of the \textit{Direction} aspect. The \checkmark and \crossmark in the parentheses are assessed by our automatic evaluation method. The \textit{Caption Generation} task instruction is discarded for simplicity.}
    \label{tab:eval_example_7}
\end{table*}
\begin{table*}[t]
    \centering
    \begin{tcolorbox}    
    \includegraphics[width=0.7\textwidth]{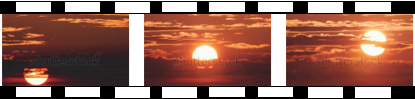} \\
    
    \textbf{Multi-Choice QA}: \\
    {\color[HTML]{EF0000} Q:} What is the primary movement of the sun?
    A. descending
    B. standing still
    C. rising \\
    {\color[HTML]{00CC00} Ground-Truth:} C. rising\\
    \hspace*{1em} {\color[HTML]{0000EF} SPHINX-v2:} C (\checkmark)\\
    \hspace*{1em} {\color[HTML]{0000EF} Video-LLaVA:} The sun is seen rising in the sky, which is the primary movement of the sun in the video. (\checkmark)\\
    \hspace*{1em} {\color[HTML]{0000EF} Video-ChatGPT:} The primary movement of the sun in the video is descending, as it is seen setting over the ocean. (\crossmark)\\
    
    \textbf{Yes/No QA}: \\
    {\color[HTML]{EF0000} Q:} Is rising the primary movement of the sun? \\
    {\color[HTML]{00CC00} Ground-Truth:} yes\\
    \hspace*{1em} {\color[HTML]{0000EF} SPHINX-v2:} No (\crossmark)\\
    \hspace*{1em} {\color[HTML]{0000EF} Video-LLaVA:} Yes. (\checkmark)\\
    \hspace*{1em} {\color[HTML]{0000EF} Video-ChatGPT:} Yes, the sun is rising in the sky. (\checkmark)\\
    
    \textbf{Caption Matching}: \\
    {\color[HTML]{EF0000} Q:} Which description is a more suitable match for the video?
    Option 1: The sun is moving downwards.
    Option 2: The sun is moving upwards. \\
    {\color[HTML]{00CC00} Ground-Truth:} Option 2: The sun is moving upwards.\\
    \hspace*{1em} {\color[HTML]{0000EF} SPHINX-v2:} The sun is setting. (\checkmark)\\
    \hspace*{1em} {\color[HTML]{0000EF} Video-LLaVA:} Option 1: The sun is moving downwards. (\crossmark)\\
    \hspace*{1em} {\color[HTML]{0000EF} Video-ChatGPT:} The sun is moving downwards in the sky. (\crossmark)\\
    
    \textbf{Caption Generation}: \\
    {\color[HTML]{00CC00} Ground-Truth:} Information B: \{'subject': 'sun', 'direction': 'rising'\}\\
    \hspace*{1em} {\color[HTML]{0000EF} SPHINX-v2:} The sun is setting in the sky. (\crossmark)\\
    \hspace*{1em} {\color[HTML]{0000EF} Video-LLaVA:} The sun is seen setting in the sky, with clouds moving around it. (\crossmark)\\
    \hspace*{1em} {\color[HTML]{0000EF} Video-ChatGPT:} The consistent information with the video is that the sun is setting, as it is described as a "sunset" in the video. The other two pieces of information, "sun rising" and "sun staying in one place," do not align with the visual content of the video, which shows the sun setting. Therefore, the caption should be "Sun setting in the sky." (\crossmark)\\
    \end{tcolorbox}
    \caption{An example of MLLM responses and evaluation results of the \textit{Direction} aspect. The \checkmark and \crossmark in the parentheses are assessed by our automatic evaluation method. The \textit{Caption Generation} task instruction is discarded for simplicity.}
    \label{tab:eval_example_8}
\end{table*}
\begin{table*}[t]
    \centering
    \begin{tcolorbox}    
    \includegraphics[width=0.7\textwidth]{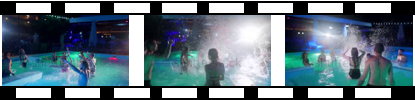} \\
    
    \textbf{Generated Caption (Video-LLaVA)}: \\
    A group of people are dancing in a pool {\color[HTML]{EF0000}while a man is cooking in the background.} \\
    \textbf{Multi-Choice Question}: \\
    What are the people doing? A. cooking B. sleeping {\color[HTML]{00BB00}C. dancing} \\
    \textbf{ChatGPT Reasoning}: \\
    The video description mentions that a group of people are dancing in a pool, while cooking is happening in the background. \\
    \textbf{ChatGPT Answer}: C. dancing \\
    \textbf{ChatGPT Rating}: \checkmark \\
    \textbf{Human Rating}: \crossmark \\
    
    \includegraphics[width=0.7\textwidth]{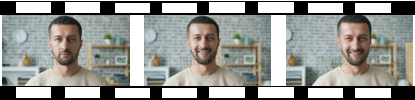} \\
    
    \textbf{Generated Caption (Video-LLaVA)}: \\
    The man in the video {\color[HTML]{EF0000}is seen smiling and then} showing a serious expression. He then smiles again. \\
    \textbf{Multi-Choice Question}: \\
    In what order does the man's expression change? A. The man doesn't change his expression B. Smiling, then showing a serious expression {\color[HTML]{00BB00}C. Showing a serious expression, then smiling} \\
    \textbf{ChatGPT Reasoning}: \\
    The video description states that the man is seen smiling and then showing a serious expression, followed by smiling again. This implies that the man's expression changes from smiling to showing a serious expression, and then back to smiling. \\
    \textbf{ChatGPT Answer}: C. Showing a serious expression, then smiling \\
    \textbf{ChatGPT Rating}: \checkmark \\
    \textbf{Human Rating}: \crossmark \\
    \end{tcolorbox}
    \caption{Examples showing our LLM-evaluation fail to detect the unsatisfactory caption in terms of hallucination. The hallucinated content in MLLM response is highlighted in {\color[HTML]{EF0000}red}. The correct answer to the multi-choice question is highlighted in {\color[HTML]{00BB00}green}.}
    \label{tab:eval_example_captioning}
\end{table*}

\end{document}